\PassOptionsToPackage{table,xcdraw}{xcolor}
\documentclass[acmsmall, screen=true]{acmart}

\AtBeginDocument{%
  \providecommand\BibTeX{{%
    \normalfont B\kern-0.5em{\scshape i\kern-0.25em b}\kern-0.8em\TeX}}}

\setcopyright{acmcopyright}
\copyrightyear{2023}
\acmYear{2023}
\acmDOI{XXXXXXX.XXXXXXX}

\acmJournal{JACM}
\acmVolume{37}
\acmNumber{4}
\acmArticle{111}
\acmMonth{8}

\usepackage{algorithmic}
\usepackage{amsmath}
\usepackage{color}
\usepackage{algorithm}
\usepackage{graphicx}

\usepackage{CJKutf8}

\usepackage{verbatim}

\usepackage{amsmath}

\usepackage{makecell}

\usepackage{graphicx}
\usepackage{footmisc}
\usepackage{subfigure}
\usepackage{url}
\usepackage{multirow}
\usepackage{diagbox}
\usepackage{bbding}

\usepackage{graphicx}

\usepackage{amsmath,amsthm}

\usepackage{amssymb,amsfonts}
\usepackage{booktabs}
\usepackage{ccaption}
\usepackage{booktabs}
\usepackage{float}
\usepackage{fancyhdr}
\usepackage{caption}
\usepackage{stfloats}
\usepackage{comment}
\graphicspath{{figures/}}
\usepackage{cuted}%flushend,
\usepackage{epstopdf}
\usepackage{booktabs}  
\usepackage{threeparttable}  
\usepackage{footnote}
\usepackage{multirow}
\usepackage{booktabs}
\usepackage{hhline}
\usepackage{diagbox}

\usepackage{stackengine}

\usepackage{booktabs}

\def\etc{ \emph{et al.}}
\def\quote#1{~\cite{#1} }
\begin{document}

\title{Artificial Intelligence for Web 3.0: A Comprehensive Survey}

\author{Meng Shen}
\authornotemark[1]
\affiliation{%
  %\department{ABC}
  \institution{Beijing Institute of Technology}
  \city{Beijing}
  \postcode{100081}
  \country{China}
  %\streetaddress{5 Zhongguancun South Street}
}
\email{shenmeng@bit.edu.cn}

\author{Zhehui Tan}
\affiliation{%
  \institution{Beijing Institute of Technology}
  \city{Beijing}
  \postcode{100081}
  \country{China}
  %\streetaddress{5 Zhongguancun South Street}
}
\email{zhehuitan@bit.edu.cn}

%Dusit Niyato is with the School of Computer Science and Engineering, Nanyang Technological University, Singapore (e-mail: dniyat

\author{Dusit Niyato}
\affiliation{%
  \institution{Nanyang Technological University}
  \city{Jurong West}
  \postcode{639798}
  \country{Singapore}
}
\email{niyato@ntu.edu.sg}

 \author{Yuzhi Liu}
\affiliation{%
  \institution{Beijing Institute of Technology}
  \city{Beijing}
  \postcode{100081}
  \country{China}
  }
\email{liuyuzhi@bit.edu.cn}

\author{Jiawen Kang}
\affiliation{%
 \institution{Guangdong University of Technology}
 \city{Guangzhou}
 \postcode{510006}
 \country{China}}
 \email{kavinkang@gdut.edu.cn}

 \author{Zehui Xiong}
\affiliation{
    \institution{Singapore University of Technology and Design}
    \city{Tampines}
    \postcode{487372}
    \country{Singapore}
}
\email{zxiong002@e.ntu.edu.sg}

\author{Liehuang Zhu}
\affiliation{%
  \institution{Beijing Institute of Technology}
  \city{Beijing}
  \postcode{100081}
  \country{China}}
 \email{liehuangz@bit.edu.cn}

 \author{Wei Wang}
 \affiliation{
    \institution{Beijing Jiaotong University}
    \city{Beijing} 
    \postcode{100044}
    \country{China}
    %\streetaddress{3 Shangyuancun}
 }
 \email{wangwei1@bjtu.edu.cn}
 
 \author{Xuemin (Sherman) Shen}
 \affiliation{
    \institution{University of Waterloo}
    \city{Waterloo}
    \state{Ontario}
    \postcode{N2L 3G1}
    \country{Canada}
 }
 \email{sshen@uwaterloo.ca}


\begin{abstract}
     %With the exponential expansion of the Internet since the 1990s, various cutting-edge technologies have emerged, enabling users to experience unparalleled interactions and engagements in cyberspace. 
     %The exponential expansion of the Internet has created many new problems, including the concentration of benefits among a select few, monopolization of platform resources, and loss of personal privacy. To solve these problems, the concept of Web 3.0 has been proposed. 
     Web 3.0 is the new generation of the Internet that is reconstructed with distributed technology, which focuses on data ownership and value expression. Also, it operates under the principle that data and digital assets should be owned and controlled by users rather than large corporations.
     In this survey, we explore the current development state of Web 3.0 and the application of AI Technology in Web 3.0. Through investigating the existing applications and components of Web 3.0, we propose an architectural framework for Web 3.0 from the perspective of ecological application scenarios. We outline and divide the ecology of Web 3.0 into four layers. The main functions of each layer are data management, value circulation, ecological governance, and application scenarios. Our investigation delves into the major challenges and issues present in each of these layers. In this context, AI has shown its strong potential to solve existing problems of Web 3.0. We illustrate the crucial role of AI in the foundation and growth of Web 3.0. We begin by providing an overview of AI, including machine learning algorithms and deep learning techniques. Then, we thoroughly analyze the current state of AI technology applications in the four layers of Web 3.0 and offer some insights into its potential future development direction.
\end{abstract}




\maketitle

\section{Introduction}
    In recent years, the concept of Web 3.0 has emerged as a response to the issues facing the current Internet landscape. These issues include the concentration of benefits among a select few, monopolization of platform resources, and loss of personal privacy. This new wave of innovation seeks to rebuild the Internet using technologies such as blockchain, cryptography, and other decentralized solutions. It aims to address the problem of data ownership and value expression on the Internet by utilizing distributed technology. This is expected to bring significant improvements to prospects in terms of technology, industry, and economy. %by promoting innovation in production methods, organizational structures, and economic models. 
    
    %The concept of Web 3.0 was first proposed by Gavin Wood, co-founder of Ethereum in 2014. He believes that Web3.0 is the blockchain-based Internet and the goal of Web3.0 is to reduce the reliance on centralized institutions by using distributed systems and cryptography

	The concept of Web 3.0 was first proposed by Gavin Wood, co-founder of Ethereum in 2014. He believes that Web 3.0 is the blockchain-based Internet and the goal of Web 3.0 is to reduce the reliance on centralized institutions. With the enrichment of Web 3.0 concepts, Web 3.0 can reconstruct the contemporary Internet from two aspects. From the perspective of data ownership, Web 3.0 is not only a readable and writable network, but also enables users to own their data and assets~\cite{web3inanutshell}. From the perspective of data management, 
    Web 3.0 is %an infrastructure owned and trusted by its users and builders, and 
    a new economic system that is jointly built and shared by users and builders~\cite{YaoQian2022web3}.

	%Despite the widespread discussions surrounding Web 3.0, a clear and universally accepted definition of the concept remains elusive. Additionally, the ecosystem of Web 3.0 has yet to be fully described. In this article, we aim to provide a fresh perspective by offering an ecological and usage-based hierarchical structure of the Web 3.0 system, and to provide a more comprehensive understanding of this emerging technology.
	
	Web 3.0 has seen significant growth and development, driven largely by the increasing interest in cryptocurrencies and blockchain technology. However, as the technology matures and evolves, it also faces several challenges to be addressed. Web 3.0 is currently facing major difficulties in terms of data supply, value circulation, and ecological governance. 
	
	\begin{enumerate}
		\item In Web 3.0 scenario, data is diverse and plentiful, stored across a distributed network, making it more challenging to manage compared to traditional centralized databases.
		\item In the value circulation system of Web 3.0, the user's identity system is complicated to use and is easily attacked by hackers. At the same time, the transaction efficiency and pricing accuracy of the circulation system needs to be further improved.
		\item The Web 3.0 ecosystem is susceptible to various operational abnormalities, such as the spread of inappropriate content by users~\cite{gangwar2021attm}, leading to a deterioration of the ecosystem's overall quality. Considering the increased autonomy and anonymity of Web 3.0's users, it can be costly and resource-intensive to detect and address.
	\end{enumerate}
	
	Driven by these problems, we look back at the establishment and development of Web 3.0. We then give a hierarchical architecture of the Web 3.0 ecology and summarize the current challenges faced by Web 3.0 in different layers. We find that the advancements in AI technology in recent years have provided new and powerful solutions to various obstacles in the development of Web 3.0. These solutions include utilizing AI for big data analysis~\cite{alessandretti2018anticipating}, AI-generated content~\cite{radford2021learning,dathathri2019plug,pawade2018story}, and detecting and classifying various forms of content such as text and video~\cite{gao2020attention}. These AI-powered technologies can be applied across different aspects and stages of Web 3.0, improving the overall functionality and user experience. %We will delve further into this topic and provide more detailed research next.

   %daozhe

    %%加入其他的综述
    %改文字表达
    Although a comprehensive survey of Web 3.0 has not yet been conducted, there have been some reviews of some components and applications of Web 3.0. For example, some surveys investigate the metaverse, an important application of Web 3.0~\cite{wang2022survey,yang2022fusing, huynh2023artificial}, focusing on its integration with blockchain and virtual reality. Other surveys focus on digital assets~\cite{mukhopadhyay2016brief}, an important component of Web 3.0, and mainly on the architecture, consensus mechanism, privacy and security of cryptocurrency. There are also surveys of decentralized networks~\cite{zarrin2021blockchain,yang2019survey} and decentralized identities~\cite{gilani2020survey}.
	Compared with the articles published on Web 3.0, we systematically introduce the development history of Web 3.0, the framework structure of Web 3.0 system, and the current application of AI in Web 3.0 ecology. The main contributions of this investigation are summarized as follows:
    % \usepackage{graphicx}
    \begin{table}[t]
    \caption{The differences between other existing works and our survey}
    \label{tab:my-table}
    \renewcommand\arraystretch{1.5}
    \resizebox{\textwidth}{!}{%
    \begin{tabular}{|c|l|l|}
    \hline
    \rowcolor[HTML]{C0C0C0} 
    {\color[HTML]{000000} Survey Paper} & \multicolumn{1}{c|}{\cellcolor[HTML]{C0C0C0}{\color[HTML]{000000} Year}} & \multicolumn{1}{c|}{\cellcolor[HTML]{C0C0C0}{\color[HTML]{000000} Description}} \\ \hline
    Wang\etc\quote{wang2022survey} & 2022 & Discussing the security and privacy threats to the Metaverse and the state-of-the-art countermeasures. \\ \hline
    Yang\etc\quote{yang2022fusing} & 2022 & Surveying how Blockchain and Artificial Intelligence (AI) fuse with the Metaverse. \\ \hline
    Huynh-The\etc\quote{huynh2023artificial} & 2023 & Exploring the role of AI in the foundation and development of the Metaverse. \\ \hline
    Mukhopadhyay\etc\quote{mukhopadhyay2016brief} & 2016 & Evaluating the strengths, weaknesses, and possible threats to the incentive mechanism of each crypto currency. \\ \hline
    Zarrin\etc\quote{zarrin2021blockchain} & 2020 & Investigating two aspects in the decentralized Internet: consensus algorithms and other cutting-edge technology. \\ \hline
    Yang\etc\quote{yang2019survey} & 2019 & Introducing the blockchain-based Internet with decentralized processing and traceable trustworthiness. \\ \hline
    Gilani\etc\quote{gilani2020survey} & 2020 & Providing the identity proofing and authentication solutions for different self-sovereign Identity solutions. \\ \hline
    Our Survey & 2023 & \begin{tabular}[c]{@{}l@{}}Proposing a hierarchical architecture of the Web 3.0 ecology from a significant amount of concrete research and \\ providing a comprehensive overview of the current state of AI in Web 3.0.\end{tabular} \\ \hline
    \end{tabular}%
    }
    \end{table}
	
	%The data supply aspect covers how data is generated, stored, and shared in a decentralized web environment. The value circulation aspect covers how value is exchanged, tracked, and accounted for, including the use of blockchain and smart contracts. The ecological governance aspect covers how the Web 3.0 ecosystem is governed and maintained, including the use of decentralized autonomous organizations (DAOs) and other forms of decentralized governance. The application scenarios aspect covers the various domains and industries in which AI has been applied in the Web 3.0 ecosystem, including finance, healthcare, and supply chain management, and the outcomes achieved through the integration of AI and Web 3.0 technology

    	\begin{itemize}
		
		\item To the best of our knowledge, we are the first to conduct a survey of the existing literature on the use of AI in Web 3.0. Through our survey, we provide a comprehensive overview of the current state of AI in Web 3.0, including the types of AI algorithms that have been applied, the results and outcomes achieved by different studies, and the key technical challenges and limitations that have been encountered. 
		\item We abstract the architecture of Web 3.0 from a great amount of concrete AI research, which presents an overview from four aspects: data management, value circulation, application scenarios, and ecological governance. We provide a comprehensive overview of the current research and challenges of the Web 3.0 ecosystem. 
		\item We not only present an overview of the existing studies in the field of AI for Web 3.0 but also provide further insights into the defects of existing studies, and discuss in detail the future research challenges and directions on AI for Web 3.0, which provides readers with possible directions for developing innovative solutions.

	\end{itemize}
	
	The rest of this paper is organized as follows. We commence with the evolution of Web technology and the classification of AI used in Web 3.0 in Section \ref{sec:back}. We present the architectural layers of the Web 3.0 ecosystem we define and formulate a case to aid understanding in Section~\ref{sec:arch}. In Section \ref{sec:Infra}-\ref{sec:app},  we respectively introduce the issues and existing AI solutions in the four layers of Web 3.0: infrastructure, interface, management, and application. In Section  \ref{sec:challenge}, we introduce the technical challenges faced by Web 3.0, and prospects for the future research direction of Web 3.0 technology. In Section \ref{sec:conclu}, we summarize the full literature. 

    \section{Background} \label{sec:back}

    In this section, we introduce the history of the World Wide Web. Since the World Wide Web was invented in 1989, it has experienced three generations, namely Web 1.0, Web 2.0, and Web 3.0. We summarize the development process and characteristics of these three generations and guide the logic behind the development of the World Wide Web. Then, we investigated the classification of AI technology and focused on what might be used in Web 3.0.

    \subsection{The Evolution of Web}

	\textbf{Web 1.0.} In 1989, Tim Berners Lee invented the World Wide Web at CERN in Switzerland, marking the beginning of the era of the Internet as an application. He then, with his team, realized the first website: http://info.cern.ch/ in the next year~\cite{berners1989tim}. In the following years, the important components that make up the web page were invented, including HTML~\cite{raggett1999html}, HTTP~\cite{fielding1999hypertext}, Browser, etc. %At the same time, W3C, the standard organization of the World Wide Web, was founded at MIT.
	Web 1.0 implements the Internet application in the era of personal computers, but most users can only use access and search functions to obtain information, rather than edit content.

	\textbf{Web 2.0.} %In 1999, Darcy DiNucci put forward the term Web 2.0 for the first time, but her interpretation of the term was different from now. 
    In 2014, the first Web 2.0 conference was hosted by O'Reilly Media and MediaLive, at which the concept of web as platform was proposed~\cite{toledano2013web}. In the same year, Mark Zuckerberg founded Facebook. In the following years, companies in different fields, including Amazon, and Google were established, which enabled users %to edit web content in different scenarios. 
    to publish, comment, like, and upload content on the network. %Since then, Web 2.0 has entered an era of rapid development until today.
	Web 2.0 realizes the network for users to edit content, making the network a huge social circle covering the whole world. However, it has spawned electronic giants, such as Facebook, and Google, causing serious privacy problems.

	\textbf{The conception of Web 3.0.} The conception of Web 3.0 is proposed to address the limitations of the current centralized web. To solve the problems of privacy leakage and monopoly of large companies, Tim Berners-Lee propose a system called Solid~\cite{BernersLee}, which is considered the prototype of Web 3.0 now. 
    It is a decentralized platform for social applications, which ensures that the user's data is managed independently of the application accessing this data. Users store their data in pods, and the application needs to comply with certain protocols for access. At the same time, distributed authentication and access control mechanisms ensure data privacy. Users' data is no longer fragmented across different platforms. They can freely migrate between different platforms and determine the access rights of services to their data.
    
    Later, Gavin Wood formalize the concept of Web 3.0, a system combining the World Wide Web with distributed technology, like blockchains and smart contracts~\cite{web3foundation}. It is generally accepted that the most important characteristic of Web 3.0 is decentralization~\cite{cao2022decentralized}.
	%Recently, Web 3.0 is also considered a new era of computing characterized by verifiable critical computing of applications and two key features are generic and measurable~\cite{liu2021make}. 
    In conclusion, we propose our definition of Web 3.0. It is the new generation of the Internet that is reconstructed with distributed technology, which focuses on data ownership and value expression. And it operates under the principle that data and digital assets should be owned and controlled by users rather than large corporations. The key features of Web 3.0 include decentralized, blockchain-based, privacy-protected, and AI-empowered.

\subsection{The Categorization of AI in Web 3.0}

    The rapid development of AI technology in recent years has brought new solutions to many challenges encountered in the development of Web 3.0, such as Big Data analytics, AI-empowered content generation, and classification of video, text, and other content. The technologies can be used in various scenarios and links of Web 3.0, such as data management, value circulation, ecological governance, etc. We conduct a comprehensive survey of AI technology and focus on those with potential applications in Web 3.0. 
	We evaluate the algorithms based on their model complexity. Based on this criterion, machine learning can be broadly categorized into two groups: traditional machine learning techniques and deep learning techniques. The main distinction between the two is the use of cascaded neural network layers in the algorithm.
	
	\emph{1) Traditional machine learning:} Traditional machine learning refers to the earlier methods and algorithms of machine learning that are based on statistical and mathematical principles, including Support Vector Machine(SVM), Decision Trees, K-Nearest Neighbors, and Naive Bayes. These methods require fewer computational resources and are easier to interpret.
	
	%\emph{k-Nearest Neighbors}~\cite{guo2003knn} is a simple and widely used machine learning algorithm. It works by finding the k closest data points to a given test instance and making a prediction based on the majority class or mean of the neighbors.  The algorithm requires a distance metric to measure the similarity between data points, such as Euclidean distance.
	 
	\emph{Support Vector Machine}~\cite{noble2006support} works by finding the best boundaries called hyperplanes to separate data into classes or predict output values. To handle non-linear relationships in the data, support vector machine uses kernel trick, which maps the input data into a higher dimensional space. Common kernel functions used in SVMs include linear kernels, polynomial kernels, Gaussian kernels, and sigmoid kernels. 
    For example, SVM can be used to improve the security of the Web 3.0 identity management system by detecting user behavior and identifying malicious users~\cite{yampolskiy2012face}.
    %For example, SVM can be used to improve the security of the Web 3.0 identity management system by detecting user behavior and identifying malicious users~\cite{yampolskiy2012face,cilia2018multi}.
	
	\emph{\textit{Na\"ive Bayes}}~\cite{webb2010naive} is a statistical learning algorithm. It is based on Bayes' theorem, which provides a way to calculate the probability of an event based on prior knowledge of the conditions likely to be associated with the event. The algorithm assumes that the features of a given data point are independent of each other. Naive Bayes can be used for data pricing~\cite{bauer2018optimal}.
	
	%\emph{Hidden Markov Model}~\cite{eddy2004hidden} is a statistical model used to describe systems with hidden states and observable outputs. It is often used to model sequences of events where the true underlying state is not directly observable, but instead, only the outputs generated by the states are visible. The goal of HMM is to estimate the most likely sequence of hidden states given a sequence of observed outputs. 
    %In Web 3.0, HMM is used to authenticate users according to their behavior habits~\cite{mahbub2019continuous}.
    %In Web 3.0, HMM is used to authenticate users according to their behavior habits~\cite{shen2017performance,mahbub2019continuous}.
	
	\emph{Decision tree}~\cite{myles2004introduction} works by building a tree model of decisions and their potential consequences. The tree consists of nodes representing tests on features and edges representing test results. %Nodes are divided into three types: root nodes, internal nodes, and leaf nodes. 
    The root node represents the first decision, internal nodes represent subsequent decisions, and leaf nodes represent the final prediction. The path from the root node to the leaf nodes represents a sequence of decisions based on the input feature values. Decision tree can be used to predict the price of digital assets~\cite{alessandretti2018anticipating}.
 
	\emph{Random Forest}~\cite{rigatti2017random} is an ensemble learning algorithm that combines multiple decision trees to make predictions for classification and regression tasks in machine learning. The algorithm trains multiple decision trees on random subsets of the training data and combines their predictions through voting or averaging to produce a final prediction. Random forest algorithm can be used for network behavior perception~\cite{yitingTIFS} in Web 3.0.

	\emph{2) Deep learning techniques:} It is a subfield of machine learning that is inspired by the structure of the brain, specifically the neural networks. It involves training artificial neural networks, which are composed of layers of interconnected nodes or artificial neurons, on a large dataset. Each layer processes the input and passes it on to the next layer until the final output is produced. The layers between the input and output layers are known as the hidden layers.
	
	\emph{Convolutional neural network}~\cite{kim2017convolutional} is a specific type of deep learning. It uses convolutional layers to learn local features and pooling layers to reduce spatial resolution. 
    %These layers work together to learn spatial hierarchies of features and perform state-of-the-art on a wide range of image recognition and processing tasks. 
    Also, CNNs have a relatively small number of parameters, making them efficient for tasks with scarce labeled data. CNN has a wide range of applications in Web 3.0, which can be used to transform the image style~\cite{Inceptionism,gatys2016image}, improve the blockchain incentive mechanism~\cite{chen2018ai}, and detect bad content~\cite{gangwar2021attm,wazir2020spectrogram,ba2021design}.
	
	\emph{Recurrent Neural Networks}~\cite{grossberg2013recurrent} is a type of neural network that specializes in processing sequential data, such as time series, text, and speech by maintaining a hidden state that allows the network to learn and maintain context from previous time steps. They are well-suited for tasks that require understanding the context of the input. %and are widely used in natural language processing, speech recognition, and other sequential data tasks. 
    There are also variants of RNNs such as LSTM and GRU that have been developed to further improve the ability of RNNs to handle long-term dependencies. RNN is used in Web 3.0 to generate content~\cite{pawade2018story}, predict the price of encrypted assets~\cite{zhao2020deep,saad2019toward}, and detect unhealthy content~\cite{chuttur2022multi}.
	
	\emph{Graph Convolutional Network}~\cite{zhang2019graph} is a deep learning architecture designed for graph-structured data. In GCN, the nodes in the graph represent entities, and the edges represent the relationships between them. %By exploiting the structure and relationships of graphs, GCNs can capture and propagate information in graphs, making them suitable for tasks such as node classification, link prediction, and graph classification. 
    The core component of GCN is the graph convolution operation, which aggregates information from neighboring nodes to generate a new representation for the current node. GCN is widely used in the protection of privacy and security in Web 3.0, such as the transaction entity recognition~\cite{abs-2104-06559}, malicious transaction identification~\cite{ChenPLLXZ21} and the perception of network behavior~\cite{jinpengTIFS}.

 	\section{Architecture of Web 3.0} \label{sec:arch}
	
	In this section, we introduce a new Web 3.0 architecture from the perspective of application scenarios and ecosystems, as shown in Fig.~\ref{fig:architecture}. In the past, the framework of Web 3.0 was often from a technical perspective, such as the Web 3.0 technology stack proposed by Gavin Wood \cite{web3foundation}. We will illustrate the rationale for dividing Web 3.0 into distinct layers and the function of each layer as well as the crucial role of AI within this context.
	
	\subsection{The Hierarchical Architecture of Web 3.0}
	
	\begin{figure*}[t]
		\centering
		\includegraphics[width=1.0\linewidth=1.0]{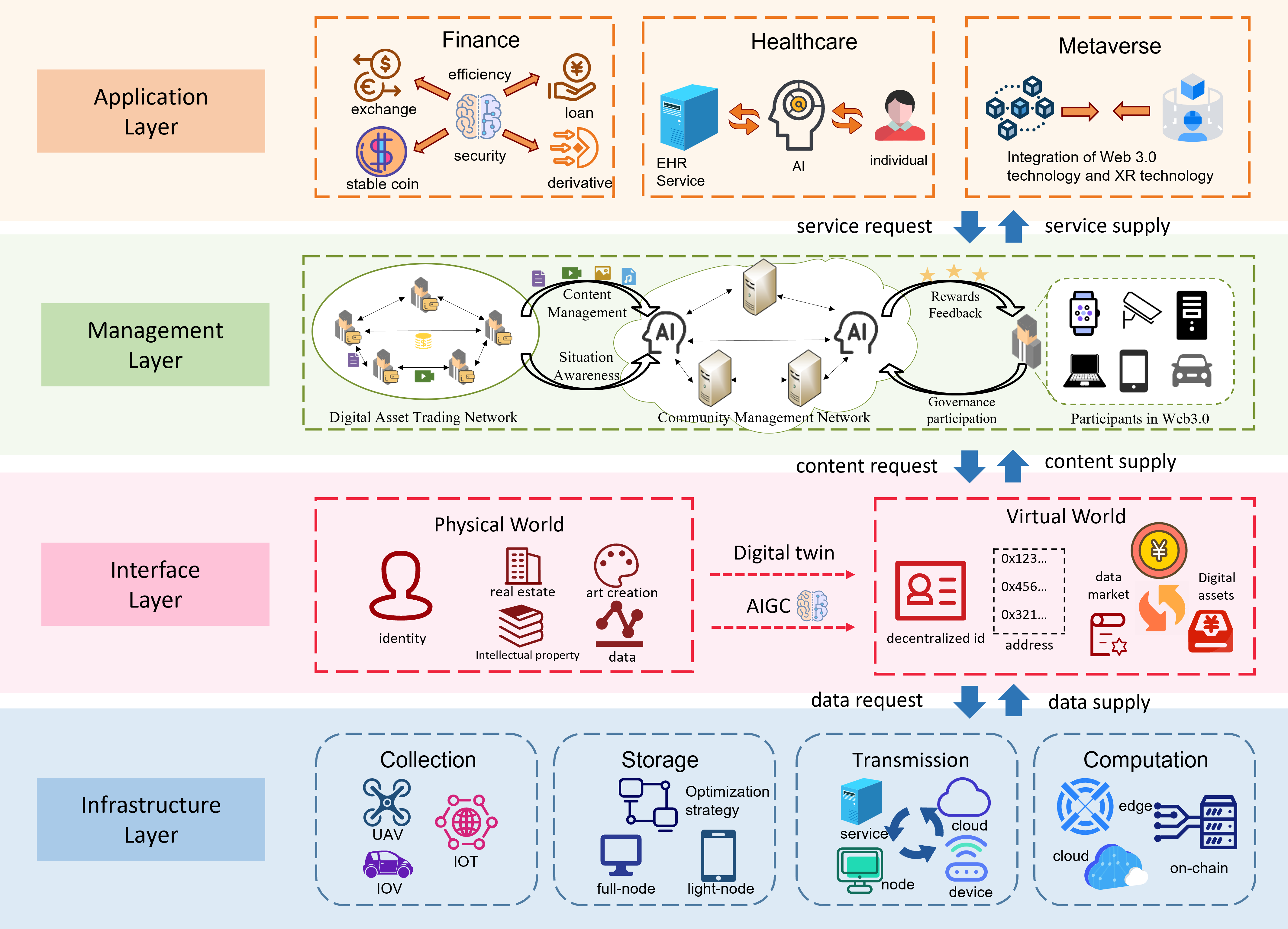}
		\caption{The Hierarchical Architecture of Web 3.0} \label{fig:architecture}
	\end{figure*}
	
	As illustrated in Fig.~\ref{fig:architecture}, Web 3.0 can be divided into four layers: infrastructure layer, interface layer, management layer, and application layer. The infrastructure layer primarily handles data management. The interface layer is responsible for mapping physical world data to the digital space. The management layer governs the ecosystem of Web 3.0. And the application layer is where actual use cases for Web 3.0 are designed and implemented. In the following section, we will explain each layer in detail.
	
	\textbf{Infrastructure layer.} It is responsible for collecting, storing, transmitting, and processing data.  With the adoption of Web 3.0 technologies, which emphasize decentralization and co-governance~\cite{qin2022web3,chen2018ai, bravo2019proof, salimitari2019ai}, the sources of data have been greatly expanded, including the use of terminal devices from the Internet of Things and real-time feedback from users. This data is transmitted to edge devices or nodes for analysis and may be stored using an on-chain or combination of on-chain and off-chain methods to ensure the transparency and effectiveness of the data. In the whole process, AI technology can be integrated into all aspects to optimize strategies, improve storage computing efficiency, improve system security, and protect privacy.
	
	\textbf{Interface layer.} It serves as a connection between the physical world and the digital world, transforming data collected from the infrastructure layer into valuable digital assets. This layer also establishes a value circulation system to optimize the utilization of data and motivate user participation.  The value of this data is determined by supply and demand, allowing users to benefit from their contributions. Unlike the Web 2.0 architecture, where private data is controlled by centralized institutions, Web 3.0 grants individuals ownership of their data, with decentralized identities responsible for protecting the privacy and access control. AI can assist in the decentralization of identity systems~\cite{yampolskiy2012face,mahbub2019continuous} and the creation of a more intelligent data market~\cite{agarwal2019marketplace}, while also safeguarding the privacy of individuals~\cite{zhou2022privacy,DBLP:journals/sensors/AliPAFMJA22,jinpengTIFS,liu2020incentive}.
	
	\textbf{Management layer.} It is responsible for the overall governance of the lower layers, including the implementation of incentives that encourage user participation\cite{momtaz2022some,xu2019incentive,zhan2019free,shinkuma2020incentive} and the review of user-generated content~\cite{moreira2020peda}. In the case of a smart city, the management layer may correct or remove erroneous or outdated traffic information uploaded by users, and detect and defend against malicious traffic attacks on the system. AI can also assist in identifying abnormal trading behavior and detecting false or inappropriate information. 
	
	\textbf{Application layer.}  It is the application system built upon the previous three layers. As an example, in the context of a smart city, this layer may include navigation apps that adjust routes in real time based on traffic conditions. Beyond the realm of smart cities, the application layer of Web 3.0 has seen significant progress in finance~\cite{9534324,DBLP:journals/corr/abs-1811-06632,10.5555/3491440.3491894}, medicine~\cite{9500397,9832978}, Metaverse~\cite{10.1145/3240508.3243653,9049708}, and healthcare~\cite{8644088}. AI technology is also heavily integrated at this level, improving user experience, enhancing privacy protection, and increasing efficiency. 
	
	\subsection{Case Study}
	
    In this section, we explore how a smart city project would function within the Web 3.0 scenario, providing a deeper understanding of the architecture of Web 3.0 that we design. 
    Smart city is a city that uses technology and data to improve the quality of life for its citizens. One important function is using sensors and information technologies to monitor and optimize traffic flow. 
    In the context of Web 3.0, This function will be implemented by the following steps. 
    
    The initial layer that plays a role is the infrastructure layer, which mainly deals with data collection, analysis, storage and other functions. To achieve real-time navigation route adjustments, it is imperative to gather current road condition information as the first step. There are two methods of obtaining this information: the first involves edge device sensors, such as cameras and Unmanned Aerial Vehicles (UAV)~\cite{liu2018blockchain}, while the second method acquiring traffic information from users and drivers. After data is collected, it will be transmitted to computing nodes and the cloud for analysis. 
    
    The interface layer receives data from the infrastructure layer, and its major function is to add value to the raw data, ensuring that it can be utilized to its fullest potential at the most relevant locations, while also incentivizing the providers of such data, thus keeping them engaged over an extended period of time~\cite{momtaz2022some}. Once these data are mapped to the Web 3.0 system, their ownership will be secured by blockchain and data confirmation technologies. Subsequently, they will be priced and circulated in the digital asset market~\cite{agarwal2019marketplace}.
    
    Then management layer will come into play, and due to the permissionless nature of Web 3.0, censorship of content will be crucial. The content review mechanism includes verifying the credibility and timeliness of transaction information, while also identifying any illegal material~\cite{moreira2020peda}, such as pornography and violence, that may be uploaded by users. 
    
    Finally, the application layer is built on the above system. Based on the services and data provided by the lower layer, developers can realize a wide range of commercial applications. In the context of smart cities, there is navigation software that can adjust routes according to real-time traffic conditions and recommendation software that can provide local service (catering, entertainment, etc.) information according to user preferences.

	% \usepackage{graphicx}
	\begin{table*}[t]
		\renewcommand\arraystretch{1.5}
		\caption{Commmon abbreviations and explanations used in this paper}
		\label{tab:my-table}
		\resizebox{\textwidth}{!}{%
			\begin{tabular}{|c|c|c|c|c|c|}
				\hline
				\rowcolor[HTML]{C0C0C0} 
				\textbf{Abbreviation} & \textbf{Explanation}                    & \textbf{Abbreviation} & \textbf{Explanation}               & \textbf{Abbreviation} & \textbf{Explanation}                 \\ \hline
				ADS                   & Authenticated Data Structure            & FCN                   & Fully Convolutional Neural Network & NLP                   & Natural Language Processing          \\ \hline
				ANN                   & Artificial Neural Network               & FL                    & Federated Learning                 & PUL                   & Positive and unlabeled learning      \\ \hline
				AoI                   & Age of Information                      & GA                    & genetic algorithm                  & QoS                   & Quality of Service                   \\ \hline
				BI                    & Bayesian inference                      & GAN                   & Generative Adversarial Networks    & RBFNN                 & radial basis function neural network \\ \hline
				CLIP                  & Contrastive Language-Image Pre-Training & GBDT                  & Gradient boosting decision trees   & RF                    & Random Forest                        \\ \hline
				CNN                   & Convolutional Neural Network            & GCN                   & Graph Convolutional Network        & RL                    & Reinforcement Learning               \\ \hline
				CNNs                  & Capsule Neural Network                  & GNN                   & Graph Neural Network               & RNN                   & Recurrent Neural Network             \\ \hline
				DAG                   & Directed Acyclic Graph                  & HAN                   & Hierarchical Attention Network     & SL                    & Supervised Learning                  \\ \hline
				DBN                   & Deep Belief Network                     & HMM                   & Hidden Markov Model                & SNN                   & Siamese neural network               \\ \hline
				DCNN                  & Deep Convolutional Neural Network       & IoT                   & Internet of Things                 & SVD                   & Singular Value Decomposition         \\ \hline
				DDPG                  & Depth Deterministic Policy Gradient     & LBF                   & Learning-based Bloom Filter        & SVM                   & Support Vector Machine               \\ \hline
				DL                    & Deep Learning                           & LSTM                  & Long short-term memory             & TPS                   & Transactions per Second              \\ \hline
				DNN                   & Deep Neural Network                     & MBP                   & Multi-armed bandit problem         & UAV                   & Unmanned Aerial Vehicles             \\ \hline
				DQN                   & Deep Q-Network                          & MC                    & Markov Chain                       & VGGNet                & Visual Geometry Group Network        \\ \hline
				DRFNet                & Dilated Residual Feature Net            & MDP                   & Markov Decision Process            & WCN                   & Wireless Communication Network       \\ \hline
				DRL                   & Deep Reinforcement Learning             & MEC                   & Moblie Edge Computing              & WSN                   & Wireless Sensor Network              \\ \hline
			\end{tabular}%
		}
	\end{table*}

\section{Infrastructure layer} \label{sec:Infra}
\begin{figure}[!htbp]
		\centering
		\includegraphics[width=1.0\linewidth=1.0]{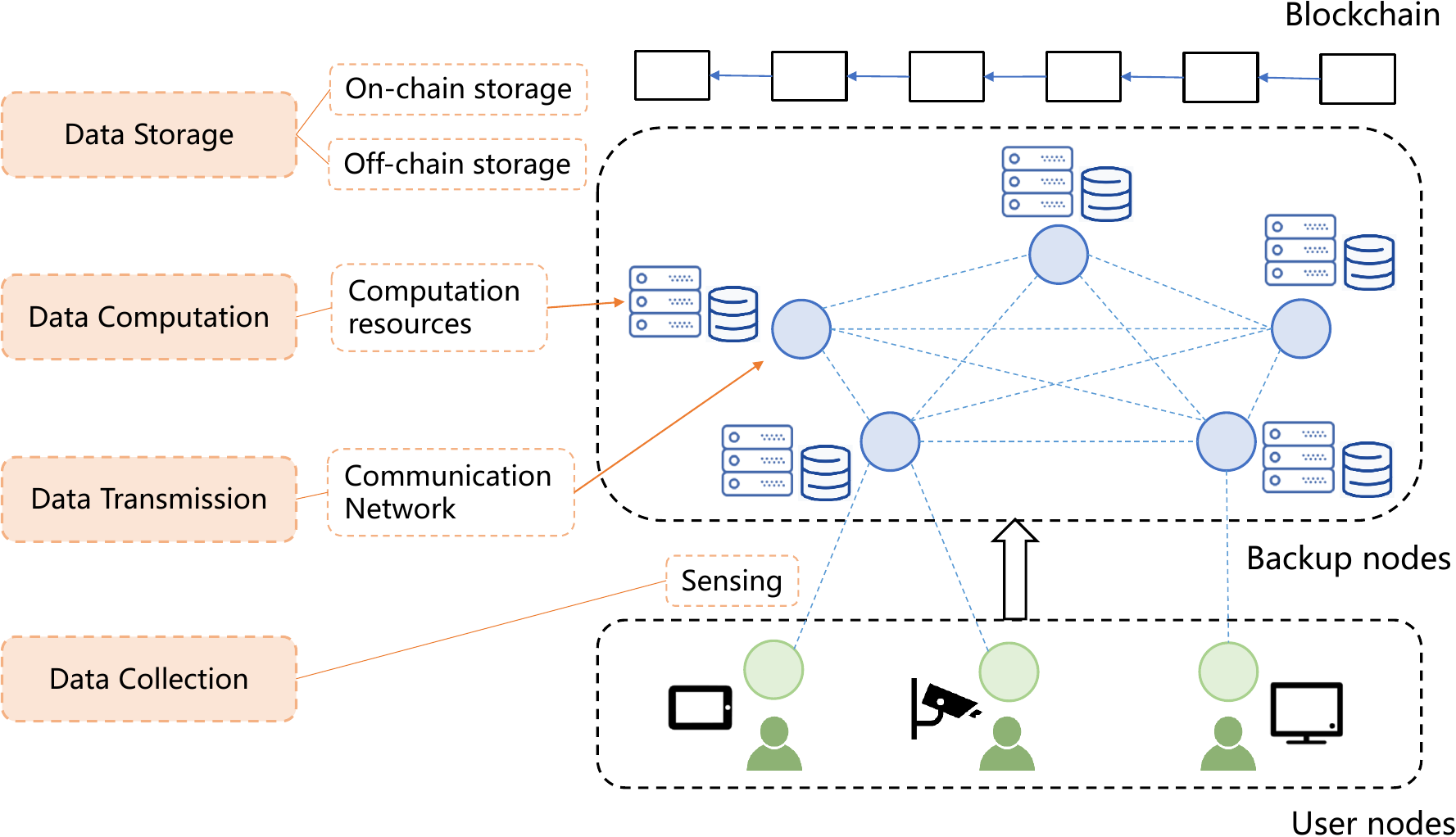}
		\caption{A common data flow process in the infrastructure layer of Web 3.0} \label{fig:infr_layer}
	\end{figure}

	The infrastructure layer of Web 3.0 is primarily responsible for data processing. which is like the foundation of a building. In Web 3.0, data is stored in a decentralized blockchain system and has various types, large quantities, and distributed storage, which is more difficult to manage than traditional databases, as shown in Fig.~\ref{fig:infr_layer}. Compared with the traditional centralized system, the Web 3.0 system has higher requirements for performance due to its transaction latency, and more evaluation indicators such as decentralization, scalability, and resource cost of blockchain-related operations which require consideration. Therefore, a specific data management solution for Web 3.0 scenarios is needed.

	\subsection{Data Collection}
	The data volume in Web 3.0 system is huge and requires many nodes and validation, which brings challenges for related studies. In Web 3.0, Internet of Things (IoT) devices play a crucial role as they provide reliable data support for web applications and services by integrating digital information between things. Therefore, Web 3.0 applications often require the support of Unmanned Aerial Vehicles (UAV)~\cite{DBLP:journals/jsac/TangLLZH22}, the Internet of vehicles\cite{ZhuangYLCR20}, etc., in which Artificial Intelligence (AI) optimizes the behavior of peers to maximize the overall system performance. At the same time, for Web 3.0 problems such as scalability and edge intelligence, the policies of the system or each node can also be optimized by AI.

	The data collected by Web 3.0 system has the characteristics of large quantity and diversity. Collection strategies should be adopted in the system to avoid unreasonable allocation of resources. AI can be used to optimize the data collection strategy, with deep reinforcement learning (DRL) being a common approach. Liu\etc\quote{liu2018blockchain}propose an efficient data collection and secure sharing scheme based on blockchain. A method based on distributed DRL is adapted to allow each mobile terminal to move to a certain location for data acquisition, maximizing the collecting rate. And according to the article, by introducing AI, energy consumption decreases from 64\% to 78\%.
	%Compared with the traditional IoT, Unmanned Aerial Vehicle (UAV) is a scenario with more flexibility and more complex behavior space, which requires a higher level of optimization. The authors in \quote{DBLP:journals/iotj/LiuGLS22}use model-based deep reinforcement learning framework to design a solution that maximizes data collection and geographic coverage while minimizing the Age of Information (AoI) of all mobile users. In this work, the index of multi-user AoI considered is consistent with the needs of Web 3.0 scenario, which also provides a reference for Web 3.0 related applications. 
	
	The convergence of IoT and Web 3.0 provides security and reliability for data collection. IoT in Web 3.0 is typically blockchain-based, and there are extra factors to be considered (e.g., blockchain-related operations). The authors in \quote{DBLP:journals/iotj/XuZYW21}propose an adaptive linear prediction algorithm. In this work, the sensed value to be uploaded is predicted and by which the actual value is replaced, thus reducing the transmission overhead. After that, Tang\etc\quote{DBLP:journals/jsac/TangLLZH22}use reinforcement learning to judge the optimal solution of each node under the PoS consensus mechanism. Different from traditional IoT, the optimal solution should be determined by taking the equity mechanism of blockchain into consideration, to achieve the global optimal solution under this mechanism.
	\begin{table*}[t]
		\centering
        \renewcommand\arraystretch{1.2}
		\caption{Research of Infrastructure Layer Based on AI}
		\resizebox{\textwidth}{!}{
			\begin{tabular}{c|c|c|c|l}
				\toprule
				\multicolumn{1}{c|}{\textbf{Subjects}} & \multicolumn{1}{c|}{\textbf{Refs}} & \textbf{AI Methods} & \textbf{Specific Scenarios} & \multicolumn{1}{c}{\textbf{Web 3.0 Tasks}} \\
				\midrule
				\multicolumn{1}{c|}{\multirow{3}[6]{*}{Data Collection}} & \cite{liu2018blockchain}     & DRL   & Industrial IoT   & Optimize collecting strategies using distributed DRL in blockchain systems \\
				\cmidrule{2-5}          & \cite{DBLP:journals/iotj/XuZYW21}     & Linear Model   & UAV-assisted IoT   & Develop prediction algorithm to save the overhead of transaction \\
				\cmidrule{2-5}          & \cite{DBLP:journals/jsac/TangLLZH22}     & DRL   & UAV-assisted IoT   & Determine the optimal strategy of edge nodes under the PoS consensus \\
				%\cmidrule{2-5} & \cite{zhou2022privacy}     & GAN   & Human data Collection & Protect the privacy of collected data in decentralized systems\\
				\midrule
				\multicolumn{1}{c|}{\multirow{4}[8]{*}{Data Storage}} & \cite{DBLP:journals/iotj/YunGC21}    & DRL   & IoT   & Determine the sharding strategy to adjust the system parameters \\
				\cmidrule{2-5} & \cite{bai2022blockchain}    & DRL   & IoT   & Incorporate the degree of decentralization for DRL training \\
				%\cmidrule{2-5}          & \cite{DBLP:journals/sensors/HuGWHL22}    & DDPG  & MEC   & Optimize the offloading and block generation strategy \\
				\cmidrule{2-5} & \cite{vairagade2022enabling}     & Genetic Algorithm    & Side chain & Determine the policy of generating the side chain \\
				\cmidrule{2-5}          & \cite{DBLP:journals/iotj/CuiSMCYZX22}     & FL    & IoT   & Enable edge intelligence for decentralized systems \\
				%\cmidrule{2-5} & \cite{wang2020fast}     & Meta-RL & Data offloading & Propose offloading method that can be used in blockchain systems \\
				\midrule
				\multicolumn{1}{c|}{\multirow{3}[6]{*}{Data Transmission}} & %\cite{DBLP:journals/access/AryaBC22}    & DBN,RL & WSN   & Develop routing protocols that can be referenced by Web 3.0 \\
				\cite{DBLP:journals/corr/abs-2205-06800}    & Multi-agent RL & WCN & Determine the distributed data transport policy \\
				\cmidrule{2-5}          & \cite{DBLP:conf/icassp/LuongABNKL19}    & DDQN  & IoT   & Derive an optimal transaction transport strategy for secondary users \\
				\cmidrule{2-5} & \cite{xu2022privacy}    & FNCF  & IoT   & Predict reliable nodes for data transmission in decentralized systems \\
				\midrule
				\multicolumn{1}{c|}{\multirow{4}[8]{*}{Data Computation}}& \cite{YangYYLZT22}    & DRL   & Collaborative computing & Determine blockchain sharding policy for the allocation of computing resources \\
				\cmidrule{2-5}          & \quote{DBLP:journals/corr/abs-2211-06861}    & DRL   & MEC   & Create incentives using ML techniques for task offloading \\
				\cmidrule{2-5} & \cite{dai2022optimizing}    & SVM,DNN    & Verifiable searching & Construct high-performanced ADS for verifiable query in blockchain \\
				\cmidrule{2-5}          & \cite{DBLP:journals/corr/abs-2101-06905}    & FL    & Decentralized FL   & Optimize the resource allocation between training and mining \\
				\bottomrule
			\end{tabular}%
		}
		\label{tab:infr_layer_tab}%
	\end{table*}%

	%Due to the high degree of flexibility and shareability of data in Web 3.0 system, security and privacy issues deserve more attention. For instance, Zhou\etc\quote{zhou2022privacy}propose a blockchain-based IoT human data (such as brain-computer interface) collection scheme to achieve the security of storing, querying and sharing personal neurophysiological data and analysis reports. The brain signals are converted into images, and the synthetic signals are generated by the generative adversarial network, and then the signals are classified by the paradigm of transfer learning.
	%In addition, Web 3.0 systems are at risk of intrusion, so intrusion detection is also a common scenario for AI. The researchers in \quote{DBLP:journals/network/RahmanTTM20}develop a machine-learning-based intrusion detection framework by training detection models, but failed to consider the data imbalance problem. To solve this, He\etc\quote{DBLP:journals/iotj/HeCTWL23}propose a CGAN-based collaborative intrusion detection framework for IoT systems, in which Long Short-Term Memory (LSTM) is applied to CGAN training to improve the performance of the networks.

	\subsection{Data Storage}
	
\begin{figure}[!htbp]
		\centering
		\includegraphics[width=1.0\linewidth=1.0]{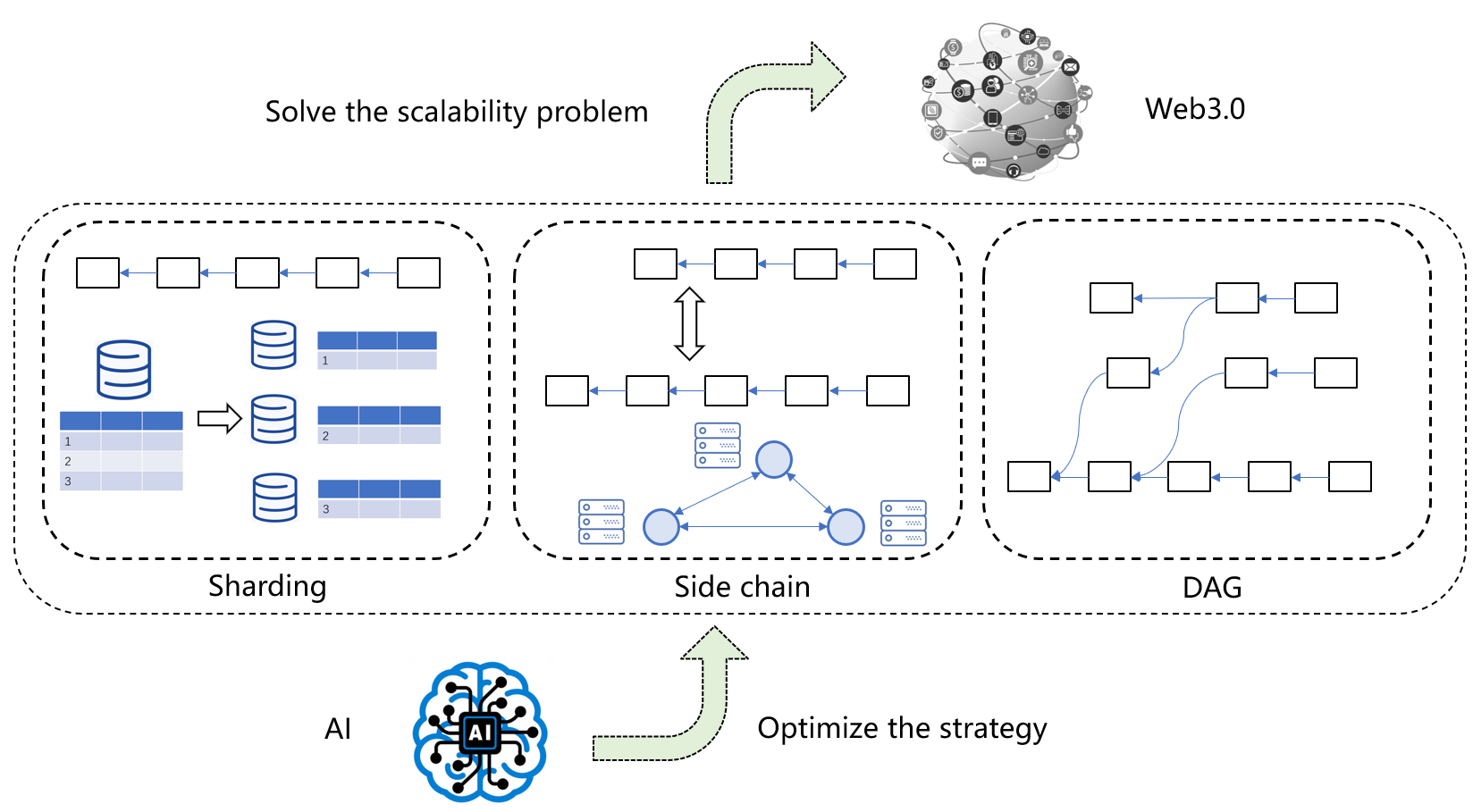}
		\caption{The Role of AI in Solving the Scalability Problem} \label{fig:scal_problem}
	\end{figure}

	Data storage in Web 3.0 is decentralized and based on blockchain. Common blockchain systems have full nodes, which store the full blockchain, and light nodes, which store only the block headers. In order to improve the scalability of Web 3.0 storage system, recent studies have proposed operation modes for Web 3.0 such as collaborative storage and blockchain sharding, in which AI can participate in the formulation of storage strategies.
	Due to the high requirements of Web 3.0 for system throughput, the storage strategy has become an important aspect, such as block generation strategy, sharding strategy, chain update strategy, etc, which provide solutions to the scalability problem, as shown in Fig.~\ref{fig:scal_problem}.
 
	\textbf{Determining the generation strategy of storage structure.}
	According to the blockchain trilemma, it is impossible for any blockchain system to take into account the following three points: scalability, decentralization and security. So the problem of optimization is an issue worth studying. The authors in\quote{DBLP:journals/iotj/YunGC21}find the sharding strategy of the blockchain and the optimal parameters of the system according to the network state through Deep Reinforcement Learning (DRL), and adaptively optimize the system throughput and security level. Also for formulating block generation policies, Bai\etc\quote{bai2022blockchain}solve the problem of quantification of the degree of decentralization of blockchain, providing conditions for system optimization. On this basis, a system optimization model based on DRL is proposed to dynamically adjust the system parameters. 
    %Aside from that, Hu\etc\quote{DBLP:journals/sensors/HuGWHL22}propose a blockchain system based on Mobile Edge Computing (MEC), which optimizes the computational offload strategy and block generation strategy. The combined performance optimization scheme is expressed using Markov Decision Process (MDP), and an algorithm based on Depth Deterministic Policy Gradient (DDPG) is designed to solve the MDP problem.
	
	Additionally, side chain is also a common solution for scalability issues and AI can help optimize the construction and update of the side chain. For example, Vairagade\etc\quote{vairagade2022enabling}use the mixed delegation practical Byzantine fault-tolerant-delegated proof of equity to update the side chain, and the continuous network information analysis is used to improve the Quality of Service (QoS). Using the modified genetic algorithm, this work optimizes the construction of side chain and solves the problem to a certain extent that assets can be stolen by malicious nodes.

	\textbf{Developing the edge storage strategies.}	
	Recent research has proposed the concept of Web 3.0 collaborative storage, in which nodes with more resources can offload part of their data to nodes with fewer resources to collaboratively use their storage resources. The data offloading technology in edge computing provides an idea for this scenario, and the technology can be transferred to this scenario. For example, Wang\etc\quote{wang2020fast}propose a task offloading method based on meta-reinforcement learning, which can quickly adapt to the new environment of a small number of gradient updates and samples. Meanwhile, mobile applications are modeled as Directed Acyclic Graphs (DAG) based on dependency. The offloading strategy is implemented by the custom sequence-to-sequence (seq2seq) neural network.
	
	As a common infrastructure for Web 3.0, IoT systems need to cache files to edge nodes to meet the requirements of large throughput. Cui\etc\quote{DBLP:journals/iotj/CuiSMCYZX22}propose a system combining IoT devices, edge nodes, and blockchain. They design an algorithm that applies the federated learning compression algorithm to the content cache to predict cached files. Each edge node uses local data to train a model, predicting popular files and improving the cache hit ratio.

	As the explosion of on-chain recorded contents and the fast-growing number of users cause increasingly unaffordable resource consumption in computing and storage in Web 3.0, Lin\etc\quote{lin2022unified}propose a unified blockchain-semantic ecosystems framework, which can convey precisely the desired meanings without consuming many resources. To achieve this, the framework utilizes dynamic sharding mechanisms to classify the same semantic demands. The dynamic sharding mechanism used in the proposed framework is adaptive and based on deep reinforcement learning (DRL), which can adjust the number of shards and their sizes based on the current workload and semantic demand patterns. By doing so, the framework can improve interaction efficiency given varied semantic demands.
 
	\subsection{Data Transmission}
	
	Web 3.0 network is a peer-to-peer structure, which is decentralized that relies on user groups to exchange information. For data transmission in the network structure of Web 3.0, AI can help solve the problem of path optimization of data transmission and formulate transmission strategies. 
	
	\textbf{Developing the transmission strategy.}	
	AI can be used to determine the transmission strategy optimization model, which helps design routing algorithms or identify the traffic changes in the network environment to dynamically adjust the data transmission volume. IoT requires better communication networks for data transfer between heterogeneous devices and an optimally Wireless Sensor Network (WSN)\cite{8954683}. %Arya\etc\quote{DBLP:journals/access/AryaBC22}develop a routing protocol based on an energy-efficient Deep Belief Network (DBN) for better transmission through the selected path. 	
    In Web 3.0 system, edge nodes tend to use distributed data transmission strategy~\cite{WuCZLSZL21}. Collin\etc\quote{DBLP:journals/corr/abs-2205-06800}study transmission control in distributed wireless communication networks from the perspective of multi-agent reinforcement learning. Each node acts as an independent reinforcement learning agent without knowledge of actions taken by other agents. %This distributed approach avoids centralized bottlenecks, and the agent can learn the cooperative transport mode to achieve efficient and fair performance.
	Considering the special uncertainty in Web 3.0 scenario, Luong\etc\quote{DBLP:conf/icassp/LuongABNKL19}use blockchain and mining pools to support IoT services based on cognitive radio networks. A deep reinforcement learning algorithm is proposed to derive the optimal transaction transport strategy for secondary users. Also, the Double Deep Q Network (DDQN) is used, which allows secondary users to learn the optimal strategy.
	
	\textbf{Enhancing the security of data transmission.}	
	From the view of data transmission, data availability is an important aspect. Typically, the edge devices need to connect to a reliable Web 3.0 node for efficient data synchronization as they complete data collection. Therefore, the reliability prediction is needed, in which AI can be used. For instance, Zheng\etc\quote{DBLP:journals/corr/abs-1910-14614}develop a framework using ML methods to predict reliable peers in blockchain systems. Then, considering the privacy issues, Xu\etc\quote{xu2022privacy}propose a personalized reliability prediction model for privacy protection through Federated Learning Neural Collaborative Filtering (FNCF) in the Internet of Things. This method allows users to protect user privacy without passing data to third parties and provides users with personalized predictions.

	\subsection{Data Computation}
	
	Web 3.0 uses distributed computing, eliminating the need for a central authority to manage and allocate resources. Resources are aggregated and optimized by the network itself, thus improving the efficiency and cost-effectiveness of resource use. Web 3.0 obtains computing capability by connecting with cloud servers, which also causes a series of challenges, such as system scalability, efficiency of computing resource integration, and privacy issues of user data. Web 3.0 data computing covers a wide range of fields, including common computing scenarios like FL, migration learning, computing resource allocation, and related operations on the database, such as data retrieval.
	
	%\textbf{Improving the performance of data computing.}	
	ML is a common data computing scenario in Web 3.0. In Web 3.0 system, distributed ML is a more common mode, which solves the problem that traditional centralized FL is susceptible to single point of failure and external attacks. Lu\etc\quote{DBLP:journals/network/LuHZMZ21}propose a distributed federated learning framework that considers resource consumption, adopting DRL-based algorithms to optimize the solution. %modeling the FL resource sharing task as a combinatorial optimization problem and adopting DRL-based algorithms to optimize the solution to this problem. 
    After that, aiming at the problem that previous work does not consider the mining overhead, Li\etc\quote{DBLP:journals/corr/abs-2101-06905}model the time of training and mining in the blockchain, and propose an optimized scheme for ML computing resource allocation.
	%Storing large amounts of data on the blockchain will cause a large overhead. Therefore, a federated learning approach can be adopted to store the learned model in the blockchain. In \cite{lu2019blockchain}, the problem of data sharing is transformed into a machine learning problem. Instead of directly sharing the original data, the original data is mapped to the corresponding mathematical model to realize the safe retrieval of data. PoQ (Proof of training Quality) consensus algorithm is proposed, which combines data model training with a consensus process to make better use of node computing resources. 
	
	Blockchain sharding is the solution to the scalability problem. However, the current throughput of sharded blockchains remains limited in terms of a high proportion of Cross-Shard Transactions (CSTs). Yang\etc\quote{YangYYLZT22}propose a cluster-based sharding strategy for collaborative computing of the IoT, in which the sharding is based on the user grouping of K-means clustering and the allocation of consensus nodes. Deep Reinforcement Learning (DRL) is combined to train the adjustment of consensus parameters to form the Markov Decision Process (MDP). %This scheme provides a solution to the problem of Web 3.0 computing resource allocation based on shared blockchain.
	%Additionally, migration learning in edge networks faces challenges, such as uncertain network environment, balance between resource consumption and model accuracy, etc. In~\cite{DBLP:conf/infocom/YuanJZLZ22}, distributed migration learning on edge networks is modeled as a long-term cost optimization problem, and a series of polynomial time algorithms are designed to solve this problem. 

	As another common scenario of data computation in decentralized systems, Collaborative Edge Computing (CEC) transfers tasks from busy edge servers to idle servers. Yet different MEC service providers have no incentive to help others. He\etc\quote{DBLP:journals/corr/abs-2211-06861}propose a collaborative mechanism whereby idle computing systems can obtain additional profits by sharing idle computing resources. They describe the social welfare maximization problem as a Markov Decision Process (MDP) and break it down into the allocation and execution of unloading tasks. %The Depth Deterministic Policy Gradient (DDPG) algorithm is used to solve the MDP problem, and the algorithm based on dynamic programming and Deep Reinforcement Learning (DRL) is used to solve the decisions. 
	
	%\textbf{Improving the performance of verifiable data retrieval.}
	Verifiability is the key advantage of Web 3.0 from the point of data retrieval. In the Web 3.0 verifiable query scenario, users need to verify the correctness and completeness of query results. To ensure the completeness of boolean queries, bloom filter may be required as a tool for constructing proof of the nonexistence of data elements. Dai\etc\quote{dai2022optimizing}propose a learning-based Bloom Filter (LBF), in which different machine learning models are used to construct multiple LBF, and the LBF with the lowest false positive rate is selected.

	%From another aspect, AI can be used to query more complex on-chain data structures. In the Web 3.0 query scenario, it may be necessary to conduct the query in the graph-based on-chain or off-chain storage structure\quote{sealey2022iota}, so GNNs are a potential method to implement such queries. ML can also be used for keyword extraction of data objects when constructing inverted index in keyword query.
	
	%Compared with the traditional centralized system, Web 3.0 query emphasizes the protection of user privacy. The authors in \cite{DBLP:journals/sensors/AliPAFMJA22} develop a secure and searchable blockchain based on deep learning as a distributed database, and propose a new cross-domain access control strategy using homomorphic encryption. Using blockchain and deep learning technology, a new privacy protection and intrusion detection framework is designed to achieve privacy protection, and achieve the best security and anonymous keyword search in the framework of the ultra-ledger structure.
	
	\subsection{Summary and Lessons Learned}
 
	The problems in the infrastructure layer can be summarized in two points, namely the scalability problem and the security problem. In the scalability problem the main role of AI is optimization, with RL being a common approach. In the security problem, the main role of AI is detection and prediction.
	AI technology can make Web 3.0 system intelligent, and significantly improve the efficiency of Web 3.0 data management in various scenarios. We have summarized the references mentioned in this chapter in Table \ref{tab:infr_layer_tab}. 	
	However, despite the work that have been done, the scalability problem and security challenges still exist in Web 3.0, which are the inherent challenges of decentralization. The current research does not completely solve these problems, and further optimization is needed.

    \section{Interface Layer} \label{sec:intface}

\begin{figure}[t]
		\centering
		\includegraphics[width=1.0\linewidth=1.0]{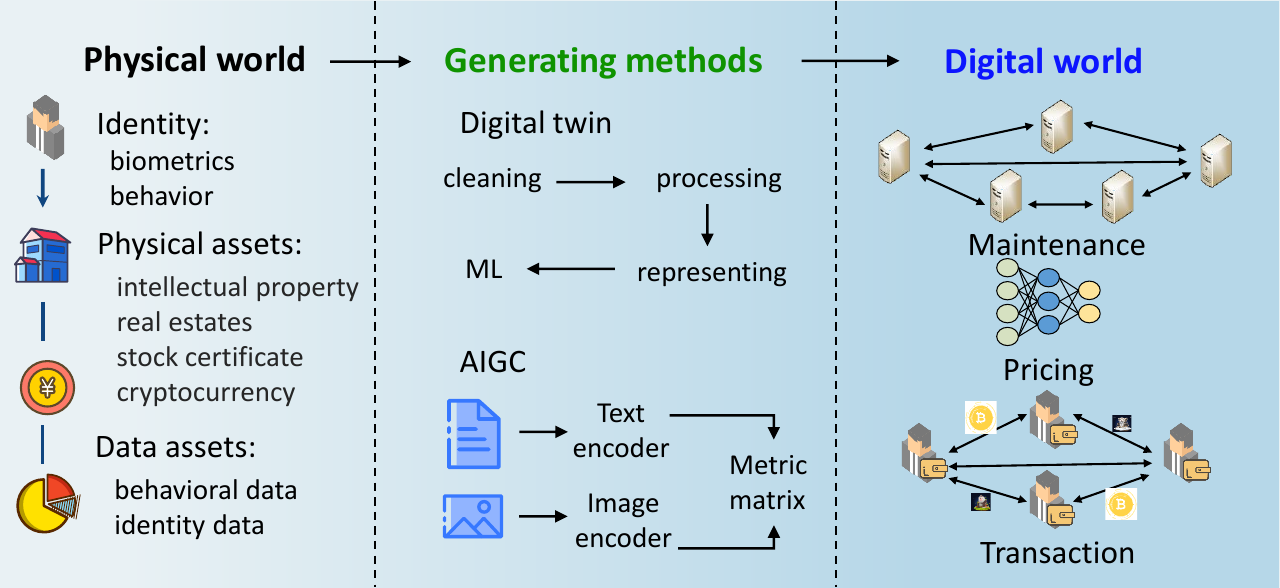}
		\caption{A common mapping model from the physical world to the digital world in the interface layer} \label{fig:interface}
	\end{figure}

    The second layer of Web 3.0 is the interface layer, which serves as a bridge between the physical and digital worlds, as shown in Fig.~\ref{fig:interface}. This layer consists of two components: digital identity and digital assets. 
    In Web 2.0, each person's digital identity on different platforms is fragmented and interoperable. At the same time, users do not have their own identities and affiliated data assets, which are monopolized by large companies. In order to solve these problems, Web 3.0 adopts a decentralized identity scheme, which ensures the user's ownership of their identity and data by storing identities on a distributed system (such as blockchain).
    Digital assets refer to valuable goods in the virtual world, including data generated by user behavior, property rights and securities that users map from the physical world to the digital world. 
    %Through Web 3.0, the ownership of these items is returned to individuals, who can obtain and maximize the value of their assets through the value circulation system.
    The introduction of digital assets means that an endogenous new equity trading market has been established in Web 3.0. This platform provides users with a value circulation system for creating, pricing, trading and consuming digital assets. 
    Through this value circulation system, digital assets can be utilized to their fullest potential by the people who need them most, and the providers can also obtain corresponding value returns.
    As a result, This digital property economy will completely change the development mode of the digital economy.

	%The second layer of Web 3.0 is the interface layer, which serves as a bridge between the physical and digital worlds, as shown in Fig.~\ref{fig:interface}. This layer is composed of two components: digital identity and digital assets. Web 3.0 utilizes a decentralized identity system, which allows users to have control over their data and resources unlike Web2, where each user's identity is fragmented and not interoperable, and ownership of the user's identity lies with monopoly organizations. 
    %Digital assets refer to valuable commodities in the virtual world, including data generated by user behavior and property rights and securities that users map from the physical world to the digital world. Through Web 3.0, the ownership of these items is returned to individuals, who can then access and maximize the value of their assets through the value circulation system.
	
	\subsection{Digital ID}

	Digital identity allows individuals to verify their identity and access online services in the digital environment.
    It is the mapping of human's real identity in the physical world to the virtual world. The decentralized identity scheme is adopted in Web 3.0.
	%Decentralized identity (DID) is a new digital identity method, which is being developed and adopted in the Web 3.0 environment. In the traditional centralized system, a single entity controls the release and management of identity information, which may lead to data leakage, personal data out of control, and identity theft. In contrast, DID is decentralized and stored on distributed ledgers (such as blockchain), enabling individuals to fully control their identity information and reduce the risk of data disclosure and identity theft.
	At present, there are mainly two ways to realize decentralized identity: W3C DID and Ethereum NFT. 
    On the one hand, the World Wide Web Consortium (W3C) has developed a set of decentralized identifier (DID) standards and protocols for creating, managing, and using DIDs~\cite{w3cdid}. 
    It includes guidelines for creating and managing DIDs, and rules for using DIDs to represent and verify identity information in a decentralized manner. 
    It can be used not only for people, but also for anything, including cars, machines, and even algorithms. 
    On the other hand, NFT can also be used as an expression of digital identity on the chain. 
    Vitalik\quote{weyl2022decentralized}demonstrate how to use the soul binding token (SBT) to code the trusted network in the economy and establish the reputation. 
    These tokens represent commitments, vouchers, and affiliations of individuals or entities and are non-transferable. 
    
	%At present, two standard approaches to decentralized digital identity and data asset governance are emerging: W3C DID and Ethereum NFT.
	%On the one hand, the World Wide Web Consortium (W3C) has developed a set of decentralized identifier (DID) standards and protocols for creating, managing, and using DIDs~\cite{w3cdid}. It includes guidelines for creating and managing DIDs, and rules for using DIDs to represent and verify identity information in a decentralized manner. It can be used not only for people, but also for anything, including cars, machines, and even algorithms.
	%On the other hand, NFT can also be used as an expression of digital identity on the chain. Vitalik\quote{weyl2022decentralized}demonstrated how to use the "soul binding" token (SBT) to code the trusted network in the real economy and establish the source and reputation. These tokens represent commitments, vouchers, and affiliations of individuals or entities and are non-transferable.

	Although the decentralized identity system has a bright future, it is currently facing challenges in identity authentication. 
    One problem is that the system may be very complex for users and need to manage complex private keys, which may be stolen or lost. 
    %This may lead to identity theft and loss of private keys, increasing the risk of using decentralized identity systems. 
    However, artificial intelligence (AI) technology can help solve these challenges by assisting with biometric authentication and behavioral authentication. By using AI in these areas, the threshold of using a decentralized identity system can be lowered, and the risk of key loss and identity theft can be reduced.

    Biometrics authentication is less prone to forgetfulness compared to knowledge-based authentication and is difficult to lose compared to token-based authentication~\cite{zhang2018advanced}. 
    Fingerprint recognition is commonly used in practice, with one typical example being the work of Svoboda\emph{et al}. They use generative convolutional networks to denoise visible details and predict missing parts of ridge patterns.
    Iris recognition is generally considered to be a secure method of biometric identification because the iris is an internal body part that is protected by the eyelid. 
    Wang\etc\quote{wang2019toward}implement the use of dilated convolutional kernels and residual learning in their deep learning framework. This method not only improves the accuracy in iris matching but also simplifies the network structure.

	However, biometric authentication has been criticized because it is vulnerable to attacks that happen after the initial authentication, while behavior recognition can provide continuous recognition. 
	Screen touch gestures are a typical way to be used in behavior recognition. Debard\etc\quote{debard2018learning}propose a method that utilizes deep neural networks features a dynamic sampling and temporal normalization component. Their approach can be adapted to different gestures, user styles and hardware variations.
    Mahbub\etc\quote{mahbub2019continuous}innovatively uses the user's habit of using the application to implement identification. They collect data from participants using smartphones, including information on device location, install, remove, or update applications, and the currently running foreground application, to implement identity authentication.
   
    Some methods have been used in some virtual systems to achieve accurate identification and access control in the virtual world. 
    Bader\etc\quote{8308432}use a combination of 3D tools and the Unreal Engine, a comprehensive collection of tools for creating 3D games and virtual spaces, to create a virtual world. To control access to this virtual world, they developed a centralized biometric authentication module using fingerprint technology. However, this method can not match the real identity with the virtual identity. Yampolskiy\etc\quote{yampolskiy2012face}propose a set of algorithms for accurately verifying and recognizing avatar faces for use in authenticating avatars within virtual worlds and for tracking a person between the real and virtual worlds in inter-reality scenarios.

	\subsection{Digital Asset}
	
	Web 3.0 is a new generation of the Internet that focuses on the circulation of value. The digital asset is its core object. Web 3.0 uses algorithms to create and distribute these assets, enabling the flow of value at a minimal cost. By utilizing blockchain technology and smart contracts, Web 3.0 allows users to create, own, and trade digital property rights on the Internet and gives users personal data ownership and the right to participate in the governance of Internet platforms and applications.
	
	There are two main types of digital assets in Web 3.0, one is the assets owned or controlled by individuals and enterprises in the form of electronic data, including digital intellectual property rights, emerging cryptocurrencies, and real-world physical assets mapped to digital assets like cars, real estate, and lands. 
    The second category is data assets, which are mainly a series of behavioral data generated by the operation of users in the digital world.  It can directly or indirectly create economic and social benefits. Web 3.0 provides a value circulation system to help digital assets circulate freely and maximize their value where they are most needed. 
    %The introduction of digital assets means that an endogenous new equity trading market has been established in Web 3.0. This platform provides users with a value circulation system for creating, pricing, trading and consuming digital assets. In this way, resources such as data can be utilized to its fullest potential at the most relevant locations.
    %As a result, The digital property economy will completely change the development mode of the digital economy.
	We divide the entire life cycle of digital assets into two stages, namely the generation of digital assets and the transaction circulation of digital assets. AI technologies all play a huge role throughout the lifecycle of digital assets. Each of these will be described below.

	\begin{table*}[t]
		\centering
        \renewcommand\arraystretch{1.2}
		\caption{Research of Digital Asset Based on AI}
        \resizebox{\textwidth}{!}{
		\begin{tabular}{c|c|c|c|l}
			\toprule
			{\textbf{Subject}} & \textbf{Ref.} & \textbf{AI Methods} & {\textbf{Specific Scenarios}} & \multicolumn{1}{c}{\textbf{Web 3.0 Task}} \\
			\midrule
			\multicolumn{1}{c|}{\multirow{3}[6]{*}{\makecell[c]{Generate digital assets\\ through digital twin}}} & \quote{wang2019digital} & CNN   & {Digital avatar} & Generating High-Fidelity 3D Avatar from a Single Image \\
			\cmidrule{2-5}          & \quote{schrotter2020digital} & DNN   & {Digital city} & Transforming 3D spatial data and city models to a virtual world \\
			\cmidrule{2-5}          & \quote{sun2021digital} & SVD DNN & {Digital avatar} & Formalize personality as digital twin models by observing users’ posting content  \\
			%\cmidrule{2-5}          & \quote{8842888} & RL    & {Improve accuracy} & Deal with the slight model or data errors  \\
			\midrule
			\multicolumn{1}{c|}{\multirow{7}[14]{*}{\makecell[c]{Generate digital assets\\ through AIGC}}} & \quote{Inceptionism} & CNN   & \multicolumn{1}{c|}{\multirow{2}[4]{*}{Style transfer}} & Produce a rather psychedelic and hallucinatory stylistic effect \\
			\cmidrule{2-3}\cmidrule{5-5}          & \quote{gatys2016image} & CNN   &       & Rendering the semantic content of an image in different styles \\
			\cmidrule{2-5}          & \quote{karras2019style} & GAN   & \multicolumn{1}{c|}{\multirow{5}[10]{*}{Generate content}} & Stochastic variation in the generated images \\
			%\cmidrule{2-3}\cmidrule{5-5}          & \quote{elgammal2017can} & GAN   &       & Generates art by looking at art and learning about style \\
			\cmidrule{2-3}\cmidrule{5-5}          & \quote{DALLE} & GAN   &       & Ggenerate images from text descriptions \\
			\cmidrule{2-3}\cmidrule{5-5}          & \quote{radford2021learning} & CLIP  &       & Maximize the similarity between real image-text pairs  \\
			\cmidrule{2-3}\cmidrule{5-5}          & \quote{dathathri2019plug} & VAE   &       & generate texts with better discourse structure and narrative flow \\
			\cmidrule{2-3}\cmidrule{5-5}          & \quote{pawade2018story} & LSTM  &       & Solve the gradient disappearance and gradient explosion in text generation \\
			\midrule
			\multicolumn{1}{c|}{\multirow{8}[16]{*}{\makecell[c]{The Circulation of\\ Digital Assets}}} & \quote{xu2016dynamic} & MBP & \multicolumn{1}{c|}{\multirow{3}[6]{*}{Data Pricing}} & Price of privacy in Personal data market   \\
			\cmidrule{2-3}\cmidrule{5-5}          & \quote{bauer2018optimal} &  BI &       & Pricing when few data points are available \\
			\cmidrule{2-3}\cmidrule{5-5}          & \quote{misra2019dynamic} & MBP &       & Decide on prices with incomplete demand information \\
			\cmidrule{2-5}          & \quote{agarwal2019marketplace} & DNN   &  {Transaction Matching} & A market mechanism to price training data and match buyers to sellers  \\
			\cmidrule{2-5}          & \quote{zhao2020deep} & LSTM  & \multicolumn{1}{c|}{\multirow{2}[4]{*}{Assets Pricing}} & High-performance model Prediction in the case of insufficient data samples \\
			%\cmidrule{2-3}\cmidrule{5-5}          & \quote{jang2017empirical} & BNN  &       & prediction of Bitcoin prices based on BNNs \\
			\cmidrule{2-3}\cmidrule{5-5}          & \quote{alessandretti2018anticipating} & GBDT &       & Comparison of trading strategies based on different neural network methods \\
			\cmidrule{2-3}\cmidrule{5-5}          & \quote{saad2019toward} & LSTM  &       & Price prediction using user behavior data \\
			\bottomrule
		\end{tabular}%
        }
		\label{tab:addlabel}%
	\end{table*}%

	\textbf{The generation of digital assets.}
	The generation of digital assets includes the mapping from the physical world to the virtual space and the generation of native virtual assets. 
    Two technologies used for this process are Digital Twin and AIGC. 
    Digital twin technology can be used in Web 3.0 to build smart cities, virtual avatars, and virtual world infrastructure~\cite{shen2021holistic}. 
    AI technology can help improve the efficiency and accuracy of digital twins. At present, the original digital assets are mainly digital collections and NFT, and the forms are pictures, texts, music, etc. 
    The traditional generation method is costly and inefficient. AIGC (AI Generated Content) can help creators to try out the inspirational scheme more efficiently and directly in the early stage, and it saves manpower to complete the details in the later stage.

	\begin{figure}[!htbp]
		\centering
		\includegraphics[width=1.0\linewidth=1.0]{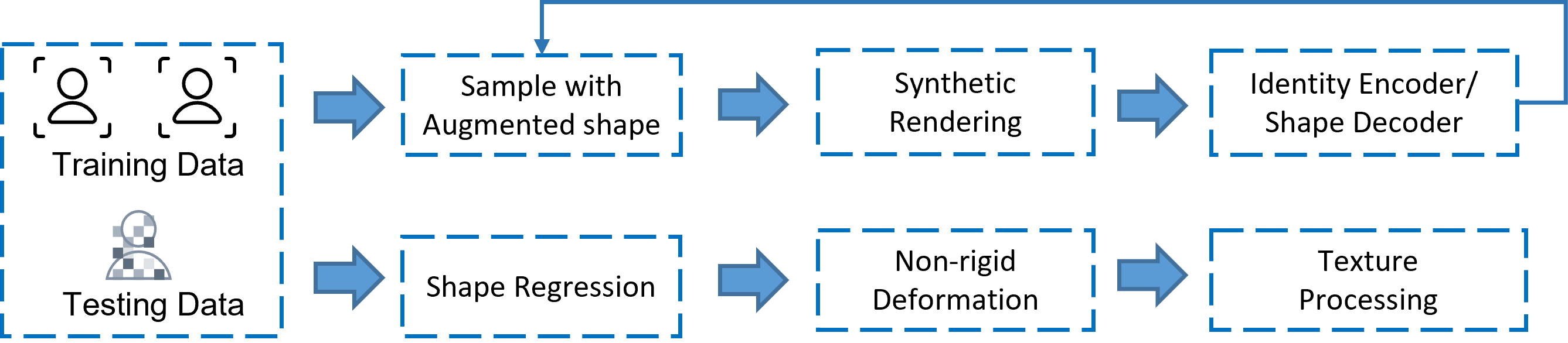}
		\caption{The method proposed in~\cite{wang2019digital}. During training, a shape regression neural network is used on photo-realistic synthetic facial images. During testing, a low polygon count shape model with a UV diffuse map generated from the projected texture.} \label{fig:digitalAssets}
	\end{figure}
	
	Digital twin is a method of generating digital assets, which refers to mapping objects in the physical world to the digital world. 
    Digital twin can help build users' digital avatars in the virtual world. 
    Wang\etc\quote{wang2019digital}develope a method for creating high-quality 3D face avatars with detailed texture maps from a single 2D image. 
    %They use a deep neural network to predict the vertex coordinates of the 3D face model from the input image, and then apply a non-rigid deformation process to refine the geometry and accurately capture facial landmarks. 
	To create more lifelike and engaging digital avatars, we should focus on not just physical characteristics, but also on developing unique personalities and preferences. Sun\etc\quote{sun2021digital}propose a method for creating digital twin models of personalities by analyzing a user's posting content and liking behavior. %The technique involves the use of a multitask learning deep neural network model to predict personality based on two types of data representation. 
	Digital twins enable the creation of a virtual replica of real-world cities. Schrotter\etc\quote{schrotter2020digital}develop a digital twin of the city of Zurich and create a virtual representation of the city by converting 3D spatial data and city models, such as buildings, bridges, and vegetation, into a virtual environment. They use machine learning techniques to predict the urban climate of the city based on current weather and air quality data. 
	
	%Despite efforts to accurately replicate the physical aspects of a system in a Digital Twin, slight discrepancies in the model or data may persist. To address any residual errors in Digital Twins, Jaensch\etc\quote{8842888}have developed an algorithm that utilizes Reinforcement Learning (RL) and feedback data from the manufacturing system. Their findings show that this approach leads to a rapid adaptation and improved performance of the autonomous system.

	%\quote{li2022big}The study aims to conduct big data analysis (BDA) on the massive data generated in the smart city Internet of things (IoT), make the smart city change to the direction of fine governance and efficient and safe data processing. 
	
	%Moya\etc\quote{moya2022digital}This novel digital twin is able, therefore, to see, to interpret what it sees—and, if necessary, to correct the model it is equipped with—and presents the resulting information in the form of augmented reality. 

	AIGC is another method of generating digital assets. It can be utilized as a tool for creating variations of images by altering their style. The first approach to receive significant attention in this field is DeepDream~\cite{Inceptionism}, a pioneering method developed by Mordvintsev. %It is originally designed to make deep convolutional neural networks more interpretable by visualizing the patterns that activated neurons the most. 
    It can create a distinctive, psychedelic visual style, leading to its use as a form of digital art.
	The separation between content and style is one of the most iconic milestones in the field of style transfer. Gatys\etc\quote{gatys2016image}first propose this idea. 
    Their algorithm can manipulate natural images by separating and recombining their content and style. With this algorithm, users can generate new, high-quality images that incorporate the content of any photograph with the style of various famous artworks.
	%daozhe
	
	One of the most significant technological advancements driving the current AI art movement is the use of Generative Adversarial Networks (GANs). The use of GANs has resulted in the generation of realistic, vivid images for various types of content, such as StyleGAN~\cite{karras2019style} and BigGAN~\cite{zhu2017unpaired}. 
    %Karras\etc\quote{karras2019style} propose a new generator architecture called StyleGAN that allows for the inclusion of stochastic variation, such as freckles and hair, in the generated images.
	Most GAN models will only learn how to generate images that look like art that already exists, and in a similar way to the NST method, this will not produce anything truly artistic or novel. %Elgammal\etc\quote{elgammal2017can}introduce aican, an AI creative adversarial network. They propose modifications to optimization criteria to enable the network to generate creative art by maximizing deviations from established styles while remaining within the artistic distribution.
	
	However, significant advancements have been made in the field of image generation from text recently. Radford\etc\quote{radford2021learning}introduce the CLIP model, which is a pre-trained model that has been trained on a large number of image-text pairs from the Internet using contrastive learning. This means that it maximizes the similarity between real image-text pairs and minimizes the similarity between mispairs. In January 2021, based on the CLIP model, DALL-E\quote{DALLE}is proposed by OpenAI, which is a 12 billion parameter neural network-based image generation system developed by OpenAI. It is trained on a dataset of text-image pairs and can generate images from textual descriptions. DALL-E can generate a wide range of images, from photorealistic to highly stylized, and can even generate images of objects and scenes that do not exist in the real world. 
	
	AI technology has also made breakthroughs in other forms of content generation in recent years, such as natural language generation(NLG). 
    With the rapid development of the deep learning neural network, the current NLG models are mainly based on deep learning neural networks and utilize a vast corpus of human-written text.
	Graves\etc\quote{graves2013generating}use recursive neural networks. However, RNN has the problem of gradient explosion and gradient disappearance. 
    To overcome these challenges, Pavade\etc\quote{pawade2018story}propose a text generation model based on LSTM. Another option is GRU, which is another extension of the standard RNN and is simpler than LSTM.  
    Many deep generation architectures use GRU to generate text~\cite{hong2018combining}.
	After that, the wide application of encoder-decoder architecture opened a new chapter. 
    Although the sequence-to-sequence model is originally developed for machine translation, it soon proved that it could improve the performance of NLG tasks. 
    Bowman\etc\quote{bowman2015generating}propose an attempt at the VAE (Variational AutoEncoder) text generation model. 
    They use recurrent neural networks to capture the general characteristics of sentences in continuous variables, such as theme and style. 
    Later, Dathathri\etc\quote{dathathri2019plug}used VAE to learn and generate texts with better discourse structure and narrative flow.

    Recently, ChatGPT~\cite{CHATGPT} has achieved great success in large-scale natural language processing models, which is a typical application of Pretrained Foundation Models (PFMs). ChatGPT is fine-tuned by the Generation Pre-training Transformer model GPT-3.5. 
    It applies convolution and recursion modules to feature extraction based on PFMs and uses autoregressive paradigms to train on large data sets mixed with text and code.  It also innovatively combines reinforcement learning from human feedback (RLHF). Due to its exceptional performance, ChatGPT has become a milestone in Natural Language Generation (NLG) and moves towards artificial general intelligence.

    %In this section, 17 papers are introduced, mainly including some milestone work in the field of generative AI and some work devoted to the generation of physical world entities using digital twins (DT) in the virtual world. Many scholars have described Web 2.0 as a network of user-generated content (UGC) We believe that Web 3.0 should be a network of user-generated content (UGC) and AI-generated content (AIGC), which is not only a mapping of the real world. More background content is needed. Work like~\cite{DALLE,CHATGPT} can play an important role. This is how the two methods DT and AIGC cooperate with each other. DT is responsible for displaying the mapping of the world, and AIGC is responsible for expanding the information in a broad sense.

	\textbf{The pricing of digital assets.}
	The second stage of the life cycle of digital assets involves the pricing and exchange of these assets. 
    AI technology is being used to create sophisticated and accurate models of digital asset prices and develop algorithms that match buyers and sellers.
    The increasing value of personal data in the era of big data has brought about a significant conflict between the exploitation of this data and the protection of individual privacy. 
    One potential solution to this issue is the development of a personal data market, but determining the appropriate price for an individual's privacy remains a challenging problem. 
    Bauer\etc\quote{bauer2018optimal}suggest that data pricing can be effectively determined using a combination of kernel regression and Bayesian inference, along with a confidence interval estimation algorithm based on the Bootstrap method. This approach is suitable for use with sparse and noisy data. However, this method is not suitable for the scenario of rapid demand change and high price sensitivity.
	
	In a price-sensitive scenario, Xu\etc\quote{xu2016dynamic}propose another method that data pricing can be approached as a reinforcement learning problem for multi-armed bandit machines. 
    %They introduce two potential learning policies for this problem: 1) estimating the expected reward of a particular price by tracking how often it is accepted by data owners, and 2) using ridge regression to estimate the reward of different prices in various contexts, taking into account the time-varying value of the data. 
    However, this method faces a unique challenge with incomplete demand information. 
	To solve this problem, Misra\etc\quote{misra2019dynamic}propose a dynamic price experimentation policy based on the extension of multiarmed bandit algorithms with microeconomic choice theory. The proposed approach uses a scalable, distribution-free algorithm to solve the resulting multiarmed bandit problem. %and is shown to be asymptotically optimal for any weakly downward-sloping demand curve through analytical proof.
	Since the data is freely replicable, the current conventional market model is not feasible for data transactions. Agarwal\etc\quote{agarwal2019marketplace}propose a new data marketplace for efficiently buying and selling training data for machine learning tasks. They make two technical contributions to this marketplace: a new concept of fairness for cooperative games involving easily replicable goods, and a mechanism for auctioning combinatorial goods that is truthful and regret-free, using Myerson's payment function and the Multiplicative Weights Algorithm.
	
	When it comes to pricing narrowly defined assets in the Web 3.0 world, there is a lot of relevant AI-based research. 
    Zhao\etc\quote{zhao2020deep}present a deep learning framework based on Long Short-term Memory Networks (LSTM) to predict short-term price movements of all cryptocurrencies.  While the models presented in this study can accurately predict the movement of Bitcoin prices, they do not provide information on the extent of the price movement.
	To evaluate the performance of different existing neural network-based trading strategies, Alessandretti\etc\quote{alessandretti2018anticipating}examine the effectiveness of three models in predicting daily cryptocurrency prices for more than 1000 currencies. 
    Two models employed gradient-boosting decision trees, while the third utilized long short-term memory (LSTM) recurrent neural networks. 
    The results revealed that all three models outperformed a baseline model using simple moving averages, with the LSTM model consistently yielding the highest return on investment. 
    %However, this method does not take into account the impact of other factors on the price.
	
	However, those cryptocurrency price prediction methods rely on the use of past price indexes to forecast future prices and do not take into account the volatile behavior of network entities that may indirectly impact the price.
    Saad\etc\quote{saad2019toward} explore features in the Bitcoin and Ethereum networks that contribute to price increases. They analyze user and network activity that has a significant impact on the prices of these cryptocurrencies and use machine learning methods to build models that predict prices.

	\subsection{Summary and Lessons Learned}

    In subsubsection entitled "The generation of digital assets.", we introduce papers mainly including some milestones in the field of generative AI and practical applications of digital twins (DT) in creating digital counterparts of physical objects. While many scholars have described Web 2.0 as a network of user-generated content (UGC), we consider Web 3.0 as a network of both UGC and AI-generated content (AIGC). Web 3.0 is not only a mapping of the real world but also needs a wealth of background information, where some approaches~\cite{DALLE,CHATGPT} can play an important role. This is how the two methods DT and AIGC. DT is responsible for displaying the mapping of the real world, and AIGC is responsible for expanding the information in a broad sense.

    Web 3.0 is also a value Internet where users can create, trade and consume digital assets in Web 3.0. In the first subsubsection, we focuse on the generation of digital assets. In the second subsubsection, seven related papers are introduced, which cover the pricing of data~\cite{bauer2018optimal,xu2016dynamic,misra2019dynamic}, the trading mechanism of data market~\cite{agarwal2019marketplace}, and the pricing and trading of encrypted assets~\cite{jang2017empirical,alessandretti2018anticipating,saad2019toward}. The objective of presenting these papers is essentially to explore one question how to establish an efficient value circulation system to maximize the use of data, assets, and other means of production. 
	%Artificial Intelligence (AI) technology has shown great promise in the realm of digital identity and digital assets. From advanced authentication methods to efficient management of digital currencies, the application of AI in these areas holds the potential to enhance security and facilitate more efficiency. However, it's important to note that the implementation of AI in digital identity and digital assets also poses certain challenges that need to be addressed. In terms of digital identity, the use of AI raises privacy concerns. Additionally, the decentralized nature of many digital assets can make them difficult to regulate, raising concerns about money laundering and other illicit activities. It is important to be aware of and address these challenges to ensure the technology is deployed responsibly and ethically.
 
    \section{Management Layer} \label{sec:manage}

	The management layer is mainly composed of the services that maintain the Web 3.0 ecosystem, including incentive mechanisms, content management, and situational awareness. The users and infrastructure in Web 3.0 ecosystem generate vast amounts of data during their daily operations, and monitoring and analyzing this data is crucial to maintaining the Web 3.0 ecosystem. In this process, AI can improve the efficiency of data analysis and provide timely and positive feedback to the community when anomalies occur in the ecosystem. In the following, a brief overview of the AI technologies that support the management of the Web 3.0 ecosystem is given.

    \subsection{Incentive Mechanism}
	
	%add
	\begin{table*}[t]
		\centering
		\caption{Research of Incentive Mechanism on AI}
        \renewcommand\arraystretch{1.2}
        \resizebox{\textwidth}{!}{
		\begin{tabular}{c|c|c|c|l}
			\toprule
			\textbf{Subjects} & \textbf{Refs.} & \textbf{AI Methods} & \textbf{Scenario-oriented} & \multicolumn{1}{c}{\textbf{Web 3.0 Tasks}} \\
			\midrule
			\multirow{4}[8]{*}{ \makecell[c]{Blockchain \\Consensus\\ Mechanism}} & \quote{chen2018ai} & CNN   & Blockchain & An AI-based super nodes selection algorithm in blockchain networks \\
			\cmidrule{2-5}          & \quote{bravo2019proof} & DL    & Blockchain & Consensus Mechanism Based on Machine Learning Competitions \\
			\cmidrule{2-5}          & \quote{salimitari2019ai} & SL    & Blockchain-based IoT networks & An Outlier-Aware Consensus Protocol \\
			%\cmidrule{2-5}          & \quote{liu2021proof} & DL    & Blockchain & Empowering neural network training with blockchains consensus \\
			%\cmidrule{2-5}          & \quote{li2022collaboration} & DL    & Blockchain & A collaboration strategy in the mining pool  \\
			\cmidrule{2-5}          & \quote{wang2022platform} & FL    & Sustainable Blockchains & A Platform-Free Proof of FL Consensus Mechanism \\
			\midrule
			%\multirow{3}[6]{*}{ \makecell[c]{Crowd\\ Sensing}} &  \quote{xu2019incentive} & DL    & MCS   & Incentive mechanism for Multiple Cooperative Tasks with Compatible Users \\
			%\cmidrule{2-5}          & \quote{zhan2019free} & DRL   & Multi-leader multi-follower MCS & Incentive mechanism for free market of multi-leader multi-follower MCS \\
			%\cmidrule{2-5}          & \quote{shinkuma2020incentive} & DL    & MCS   & Incentive Mechanism for MCS in Spatial Information Prediction Using ML \\
			%\cmidrule{2-5}          & \quote{liu2020incentive} & DRL   & Multi-leader multi-follower MCS & An Incentive Mechanism for Privacy-Preserving Crowdsensing via DRL 
		  %\\
			%\midrule
			\multirow{4}[8]{*}{\makecell[c]{ Federated\\ Learning}} & \quote{jiao2020toward} & DRL and GNN & Wireless Federated Learning  & Toward an Automated Auction Framework for Wireless FL Services Market \\
			\cmidrule{2-5}          & \quote{zhan2020incentive} & DRL   & Edge ML & An Incentive Mechanism Design based DRL for Efficient Edge Learning  \\
			\cmidrule{2-5}          & \quote{zhan2020learning} & DRL   &  {/}    & A Learning-Based Incentive Mechanism for Federated Learning \\
			%\cmidrule{2-5}          & \quote{zhao2021efficient} & RL    & Horizontal Federated Learning & Efficient Client Contribution Evaluation for Horizontal Federated Learning \\
			\cmidrule{2-5}          & \quote{xu2022fair} & Clustering    & BFL   & A flexible and Incentive Redesign for BFL \\
			%\midrule
			%\multirow{2}[4]{*}{\makecell[c]{Computation \\offloading }} & \quote{xu2021privacy} & RL    & IoT-edge computing  & Privacy-preserving incentive mechanism for IoT-edge computing market \\
			%\cmidrule{2-5}          & \quote{wang2023incentive} & DRL   & IoT   & Blockchain-Assisted Intelligent Edge Caching and Computation Offloading \\
			\bottomrule
		\end{tabular}%
        }
		\label{tab:addlabel}%
	\end{table*}%

\begin{figure}[htbp]
		\centering
		\includegraphics[width=1.0\linewidth=1.0]{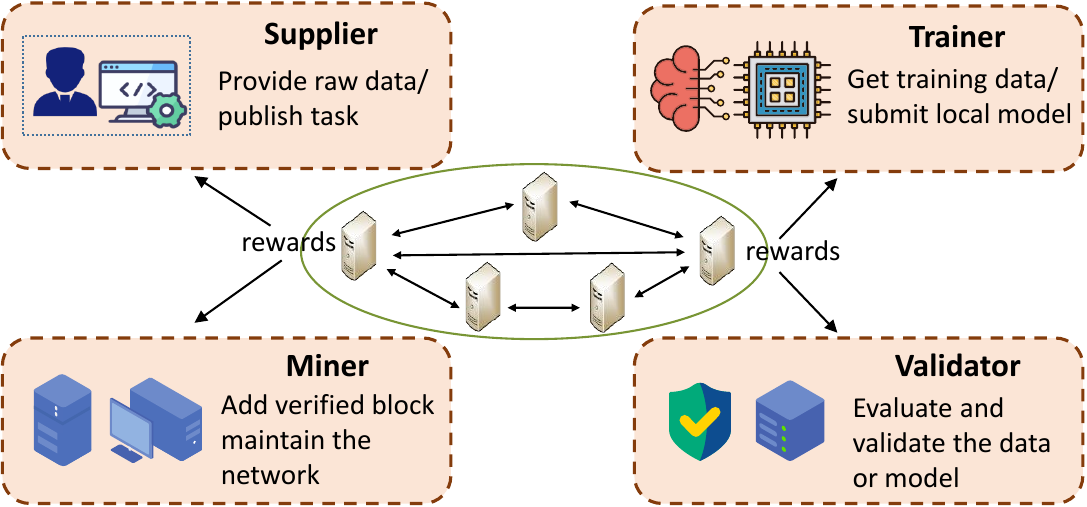}
		\caption{A typical AI-assisted incentive mechanism structure in the management layer} \label{fig:IncentiveMechanism}
	\end{figure}

    The Web 3.0 ecosystem needs to encourage users to participate in community affairs, such as encouraging users to participate in community governance, community decision-making, encouraging blockchain node consensus, node computing, and so on. Incentive mechanisms involve many aspects. According to the knowledge scenario, we focus on the incentive of blockchain consensus mechanism and federated learning scenario. As shown in the Fig.~\ref{fig:IncentiveMechanism}, it is a scenario diagram of a common incentive mechanism. 

    \textbf{Blockchain consensus mechanism.} The consensus mechanism is an incentive mechanism to encourage nodes to calculate and reach data consensus. Chen\etc\quote{chen2018ai}propose a novel node selection algorithm based on AI technology, which uses almost complementary information of each node and relies on a specially designed convolutional neural network to reach a consensus. To ensure the decentralization and security of the network, the dynamic threshold method is used to obtain super nodes and random nodes.
	
	To reduce the computational waste involved in hash-based problems, several papers discuss the possible solutions that miners' computing power will be used for relatively useful work, such as solving machine learning tasks. Bravo\etc\quote{bravo2019proof}introduce WekaCoin, which is a point-to-point cryptocurrency based on a new distributed consensus protocol called Proof-of-Learning (POLE).  Proof-of-Learning realizes distributed consensus by ranking machine learning systems for a given task.  Miners' computing ability can also be used to solve deep learning training tasks, such as Proof of Federated Learning~\cite{wang2022platform}. %Neural Architecture Search (NAS) has become popular recently because it can help researchers design deep learning models automatically. NAS needs extremely high computing power. With the help of the new consensus, the computing power of miners in blockchain services can be used to accelerate NAS. Li\etc\quote{li2022collaboration}are the first to present a mining pool solution for novel consensuses based on deep learning. The mining pool manager divides the entire search space into subspaces, and all miners are arranged to cooperate in processing NAS tasks in the allocated subspaces. The performance of this parallel computing mining pool is more competitive than that of a single miner because it is more reliable and has higher performance.	
	To reach a secure and robust consensus in the blockchain-based IoT networks, Salimitari\etc\quote{salimitari2019ai}use machine learning to propose a new framework. They introduce an AI-enabled blockchain (AIBC) with a 2-step consensus protocol, which uses an outlier detection algorithm to reach consensus in the IoT network implemented on the hyper ledger fabric platform.

	\textbf{Federated learning.} Incentive mechanism is the key design element of the new federated learning system, because: (i) participating in the federated learning will lead to the consumption of computing resources, use of network bandwidth and shorten the battery life of customers, and enough rewards can encourage them to tolerate these costs and make contributions; (ii) The worker thread in the federated learning is independent, and only its owner can determine when, where and how to participate in the federated learning. Through different incentive mechanisms, customers will implement different training strategies, which will affect the performance of the final machine learning model. In the federated learning system, the incentive mechanism has two main challenges: (i) how to evaluate the contribution of each customer, and (ii) how to recruit and retain more customers. AI technology can help solve these two challenges.
	
	For trading federated learning services in wireless environments to encourage data owners to participate in federal learning, Jiao\etc\quote{jiao2020toward}novelly develop an automated deep reinforcement learning-based auction mechanism which is integrated with the Graph Neural Network (GNN). The proposed auction mechanisms can help the FL platform make practical trading strategies to efficiently coordinate data owners to invest their data and computing resources in federated learning while optimizing the social welfare of the FL services market. 
	%To optimize Federated Learning on Non-ID Data, wang\etc\quote{wang2020optimizing}propose Favor, which is an experience-driven control framework. Based on reinforcement learning, Favor intelligently selects client devices to participate in each round of federated learning to counterbalance the bias introduced by non-IID data and to speed up convergence. They also propose a mechanism based on deep Q-learning, which learns to select some devices in each communication round to maximize rewards, encourage the improvement of validation accuracy and punish the use of more communication rounds.
	In Edge Learning, the existing work mainly focuses on the design of efficient learning algorithms, and few works focus on the design of incentive mechanisms with heterogeneous edge nodes (ENs) and network bandwidth uncertainty. Zhan\etc\quote{zhan2020incentive}propose an edge learning incentive mechanism based on Deep Reinforcement Learning (DRL), which can effectively learn the optimal pricing strategy of aggregators without knowing any prior information of ENs in dynamic networks. 
	
	In Federated Learning (FL), it is important to measure the contribution of each federal participant fairly and accurately~\cite{zhao2021efficient,xu2022fair}. Such quantification provides a reasonable metric for allocating rewards among federated clients and helps to find malicious participants who may poison the global model. The previous contribution measurement method is based on the enumeration of possible combinations of federated participants. Their calculation cost increases sharply with the increase in the number of participants or feature dimensions, making them unsuitable for the actual situation. Zhao\etc\quote{zhao2021efficient}propose an integrated contribution evaluation method F-RCCE based on reinforcement learning, which can accurately evaluate the contribution of each customer's gradient. As the number of clients increases, its time cost almost remains the same.
	
	%\textbf{Computing offloading in edge computing.} Computing offloading is a promising solution for resource-limited IoT devices to complete computing-intensive tasks. AI technology can help in the safety\quote{xu2021privacy}and performance\quote{wang2023incentive}of incentive schemes of computing offloading. Wang\etc\quote{wang2023incentive}propose an incentive-aware blockchain-assisted intelligent edge caching and computation offloading scheme to jointly optimize offloading and caching decisions as well as computing and communication resource allocation, to minimize the total cost of edge nodes (ENs) tasks completion. %A blockchain incentive and contribution co-aware federated deep reinforcement learning algorithm are designed to solve this optimization problem.

	\subsection{Content Management}

    	%add
	\begin{table*}[t]
		\centering
        \renewcommand\arraystretch{1.2}
		\caption{Research of Content Management on AI}
        \resizebox{\textwidth}{!}{
		\begin{tabular}{c|c|c|c|l}
			\toprule
			\multicolumn{1}{c|}{\textbf{Subjects}} & \multicolumn{1}{c|}{\textbf{Refs.}} & \textbf{AI Methods} &\textbf{Web 3.0 Tasks} & \multicolumn{1}{c}{\textbf{Limits}} \\
			\midrule
			\multirow{4}[8]{*}{\makecell[c]{Bad content \\detection}} & \quote{gao2020attention} & GAN   & Movie review spam detection & Poor generality: set feature words manually  \\
			%\cmidrule{2-5}          & \quote{cheng2019hierarchical} & HAN   & Cyberbullying detection & Time series analysis and Time series forecasting \\
			\cmidrule{2-5}          & \quote{moreira2020peda} & CNN   & \multirow{2}[4]{*}{Pornography image detection} & Model depends on training dataset \\
			\cmidrule{2-3}\cmidrule{5-5}          & \quote{pandey2021device} & DL    &       & Not applicable to pictures with low resolution \\
			\cmidrule{2-5}          & \quote{gangwar2021attm} & CNN   & Child sex abuse detection. & The age group detection technology is not mature \\
			%\cmidrule{2-5}          & \quote{wazir2020spectrogram} & CNN and TL & \multirow{2}[4]{*}{Inappropriate speech content detection} & Model depends on a specific data set \\
			%\cmidrule{2-3}\cmidrule{5-5}          & \quote{ba2021design} & CNN and RNN &       & Model performance decreases on classes with fewer samples \\
			%\cmidrule{2-5}          & \quote{cifuentes2022survey} & DL    & Pornographic video detection &  \\  
			%\cmidrule{2-5}          & \quote{chuttur2022multi} & LSTM and VGGNet  & Children's unhealthy content detection & Expand training data set and preprocess image frame  \\
			\midrule
			\multirow{4}[8]{*}{\makecell[c]{Deepfake \\detection}} & \quote{li2020fighting} & CNN   & Detect the deepfake image and video & Compression level and resolution are ignored \\
			\cmidrule{2-5}          & \quote{zhao2021learning} & CNN and CL  & Detect the deepfake image & The model ignores low-quality data \\
			\cmidrule{2-5}          & \quote{mittal2020emotions} & CNN and SNN & \multirow{2}[4]{*}{Detect the deepfake video} & Not applicable to multi-person video \\
			%\cmidrule{2-3}\cmidrule{5-5}          & \quote{tariq2021one} & CNN   &       & Different levels of video pressure are ignored \\
			\cmidrule{2-3}\cmidrule{5-5}          & \quote{hu2021dynamic} & CNN   &       & Poor versatility and low quality video data is ignored \\
			%\cmidrule{2-3}\cmidrule{5-5}          & \quote{gu2022region} & CNN   &       & Need to optimize the selection of time base quantity \\
			\bottomrule
		\end{tabular}%
        }
		\label{tab:addlabel}%
	\end{table*}%
 
	%\begin{figure}[htbp]
		%\centering
		%\includegraphics[width=1.0\linewidth=1.0]{images/contentManagement.pdf}
		%\caption{A typical AI-assisted content moderation workflow in Web 3.0} \label{fig:contentManagement}
	%\end{figure}
	%add
	The Web 3.0 ecosystem encourages users to create content. Users can create their articles, pictures, and videos. However, malicious content will have a serious impact on the Web 3.0 ecosystem, and AI plays an important role in the content management of the Web 3.0 ecosystem. We divide the bad content of Web 3.0 into two categories: bad content (real) and deepfake content (fake).
	
	\textbf{Bad content detection.} We divide bad content into bad text, bad image, and bad video according to the form of content. AI technology can provide great help in bad content detection.
	Bad text detection includes junk comments detection, online cyberbullying detection, etc. %Hate speech is the speech in that one person or group attacks another based on gender, race, religion, disability, or sexual orientation. %Jahan\etc\quote{jahan2021systematic}systematically review the literature in the field of hate speech detection and tracking, focusing on natural language processing and deep learning technologies. 
    Junk comments will mislead users and affect their trust in online comments. Many papers use methods based on AI technology to detect junk comments~\cite{gao2020attention,jian2022fake}. Jian\etc\quote{jian2022fake}propose a multimodal fake review detection model BAM (BERT + Attention + MLP), which uses neural networks as well as multimodal fusion technology to realize the recognition of fake reviews.  
	%As one of the urgent risks faced by young people, cyberbullying has aroused great concern in society. To solve the problem, Cheng\etc\quote{cheng2019hierarchical}propose a Hierarchical Attention Network for Cyberbullying Detection (HANCD) framework. 
	
	In terms of bad image detection, we focus on pornography detection and child sexual abuse detection. For pornography detection, Moreira\etc\quote{moreira2020peda}contribute a pornographic dataset (PEDA 376K) and propose a deep learning architecture for training on this dataset, which has excellent performance. For detecting nudity and semi-nudity contents, Pandey\etc\quote{pandey2021device}propose a deep learning solution ensemble containing MobileNetV3 classifier and SSD with MobileNetV3 feature extractor. SSD detects unsafe body parts while also providing a human-localized portion. To solve the Child Sexual Abuse (CSA) problem, Gangwar\etc\quote{gangwar2021attm}propose a deep CNN architecture with a novel attention mechanism and metric learning, denoted as AttM-CNN. 
	%Filtering audio content has become one of the major concerns of society. Many papers use methods based on AI to detect inappropriate speech content~\cite{wazir2020spectrogram,ba2021design}. For instance, Wazir\etc\quote{wazir2020spectrogram}propose an intelligent model to detect foul language censorship through automatic and robust detection of the network. %Ba\etc\quote{ba2021design}propose an intelligent system that uses advanced depth convolution neural networks and recurrent neural networks with long and short-term memory cells to detect coarse language censorship through mechanized and strong detection methods.   
	For violence detection in the video, Wu\etc\quote{wu2020not}first release a large-scale multi-scene dataset called XD Violence and propose a neural network with three parallel branches to capture different relationships between video clips and integrate features. %Aiming at the problem that many prior works overlook the modality heterogeneousness over the weakly-supervised setting, Yu\etc\quote{yu2022modality}propose a modality-aware contrastive instance learning with self-distillation (MACIL-SD) strategy considering the modality asynchrony and undifferentiated instances phenomena of the multiple instance learning (MIL) procedures.
	
	%For pornographic detection, Cifuentes\etc\quote{cifuentes2022survey}review the different strategies available for video pornography detection in the literature and identifies research gaps. Their survey shows that compared with other traditional detection strategies, the technology based on deep learning can more accurately detect videos with explicit pornographic content. For content filtering, Chuttur\etc\quote{chuttur2022multi}suggest evaluating the effectiveness of classifying videos as suitable or unsuitable for children using the actual image of cartoon characters and the language used in cartoons. They achieve this by developing a multimodal classifier that utilizes the network of LSTM for text analysis and VGGNet for image analysis.

    \textbf{Deepfake detection.} Deepfake content mainly includes images and videos. Artificial intelligence technology plays an important role in deepfake detection.
    The first category is deepfake image detection. An effective solution is to use image inconsistency to detect deepfake images. Li\etc\quote{li2020fighting}creatively propose a new Patch and Pair Convolutional Neural Network (PPCNN) architecture to detect deepfakes. They construct a dual-branch learning framework, which is the first to learn the difference between real and false face patches and the second to capture the inconsistency between the facial and the nonfacial region. The results of the two branches are combined when making a global decision. Zhao\etc\quote{zhao2021learning}propose a new novel representation learning approach, called Pair-wise Aelf-consistency Learning (PCL) to detect deepfake images using the cue of the source feature inconsistency within the forged image.  
	
    \begin{figure}[htbp]
		\centering
		\includegraphics[width=1.0\linewidth=1.0]{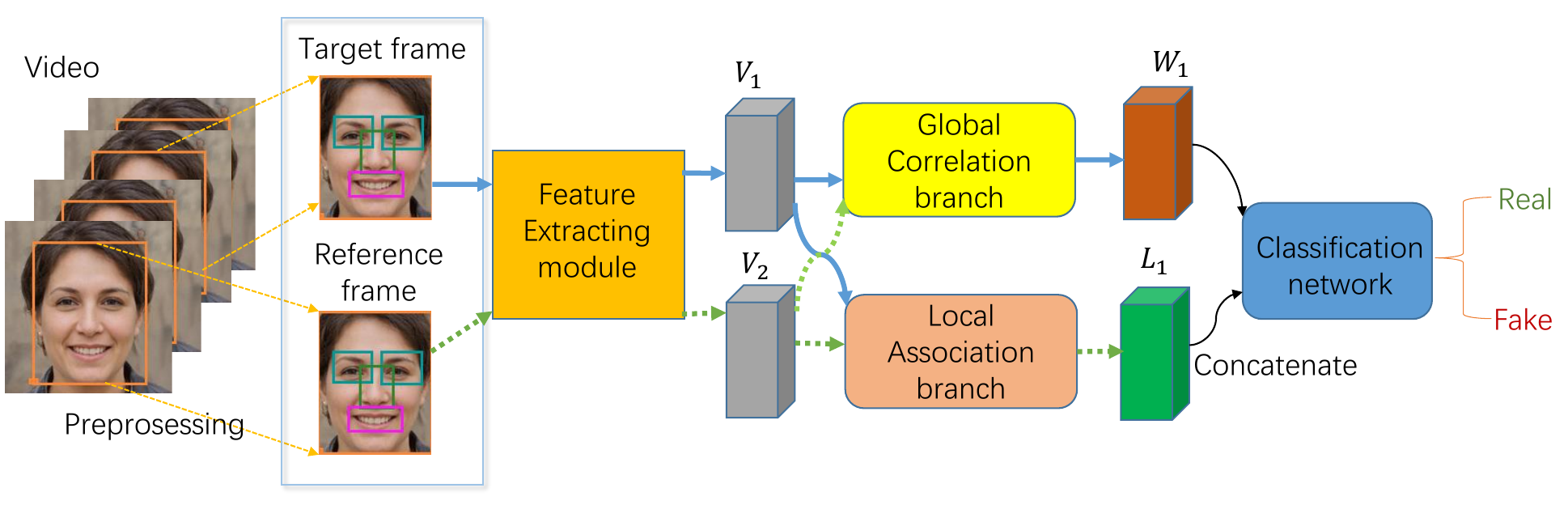}
		\caption{Technology roadmap of deepfake video detection. Utilizing the inconsistency information between adjacent frames to detect deepfake video\quote{hu2021dynamic}.} \label{fig:roadmap-b}
	\end{figure}

	%The second category is deepfake audio detection. Wang\etc\quote{wang2020deepsonar}propose a new method named DeepSonar, which is based on monitoring neuron behavior of the speaker recognition (SR) system, namely, the deep neural network (DNN), to discern AI-synthesized fake voices. There are two excellent methods~\cite{martin2022vicomtech,yan2022audio} for deepfake audio detection in ADD 2022: The First Audio Deep Synthesis Detection Challenge~\cite{yi2022add}. Martin\etc\quote{martin2022vicomtech}propose a method based on the combination of pre-trained wav2vec2 feature extractor and downstream classifier to detect spoofed audio. This method uses contextualized speech representations of different converter layers to completely capture the discriminant information. In addition, the classification model uses different data enhancement technologies to adapt to application scenarios. The audio synthesis detection system was evaluated in the ASVspool 2021 and ADD 2022 challenges, demonstrating its robustness and good performance in such challenging environments as telephone and audio codec systems, noisy audio, and partial deepfakes. Yan\etc\quote{yan2022audio}describe the best system and methodology for ADD 2022.
	
	%add
	
	The second category is deepfake video detection. Audiovisual joint detection is a popular method of deepfake video detection. Mittal\etc\quote{mittal2020emotions}propose a new method to detect any forgery or change in the input video using audio (speech) and video (face) modes and the perceptible emotional features extracted from these two modes. To simulate such multimodal features and perceived emotions, the learning method uses a Siamese network-based architecture. %Zhou\etc\quote{zhou2021joint}propose a new visual/auditory deepfake joint detection task and show that using the intrinsic synchronization between visual and auditory modes can facilitate deepfake detection. 
 
	%\begin{figure}[htbp]
		%\centering
		%\includegraphics[width=1.0\linewidth=1.0]{images/Content management technical route 2.pdf}
		%\caption{Technology roadmap of deepfake video detection. Audiovisual joint detection\quote{zhou2021joint}.} \label{fig:roadmap-a}
	%\end{figure}
	
	Another popular way to detect deepfake video is by utilizing the inconsistency information between adjacent frames~\cite{hu2021dynamic,gu2022delving}. Hu\etc\quote{hu2021dynamic}propose a new Dynamic Inconsistency-aware Network (DIANet) for DeepFake video detection by utilizing the inconsistency information between adjacent frames. As shown in Fig.~\ref{fig:roadmap-b}, DIANet consists of three modules: Feature Extraction Module, Cross-Reference Module (CRM), and Classification Network. DIANet takes a pair of frames as input and obtains their feature representation through the Feature Extraction module. Then the proposed CRM is used to capture the global and local inconsistencies between adjacent frames. Finally, the global and local inter-frame inconsistencies are combined and sent to the classification network. The model generalizes well on videos of low quality and unseen manipulation techniques.

	%Frame inference is also a method for detecting deepfake video. Hu\etc\quote{hu2022finfer}propose a frame inference-based detection framework (FInfer) to solve the problem of high visual quality Deepfake detection. Specifically, first, learn the reference representation of the face of the current and future frames. Then, the facial representation of the current frame is used to predict the facial representation of the future frame using the autoregressive model. Finally, a representation of prediction loss is designed to maximize the discriminability between a real video and a fake video. This method is promising in the detection performance, detection efficiency, and cross-dataset detection performance of high visual quality Deepcake video.	
	
	At present, most of the advanced algorithms are trained to detect specific fake methods. Therefore, these methods show poor generalization in different types of face operations from face swapping to facial reenactment. Cozzolino\etc\quote{cozzolino2021id}propose a new method that is introduced to learn the temporal facial features of how a person moves when talking through measurement learning combined with antagonistic training strategies. The advantage is that only real video training is required. In addition, the use of advanced semantic features makes it robust to extensive and destructive post-processing forms.

    \subsection{Situation Awareness}
    The trading of digital assets in Web 3.0 depends on the safe and stable cyberspace security environment, so situation awareness of network security is essential. Artificial intelligence plays a very important role in network situation awareness, which is mainly reflected in three aspects: transaction entity recognition, malicious transaction identification and network behavior recognition.
	
	%\begin{figure}[htbp]
		%\small
		%\centering
		%\includegraphics[width=\linewidth]{images/fig1.pdf}
		%\caption{Role of situational awareness module in the management layer}\label{situawar}
	%\end{figure}
	
	\textbf{Transaction entity Recognition}. Users in Web 3.0 ecosystem trade digital assets anonymously, and transaction entity identification aims to determine which addresses belong to the same entity from the vast amount of transaction records. Due to the lack of labeled datasets, unsupervised learning methods can effectively assist researchers in identifying user entities.
	
	The most common way to de-anonymize transactions is to use heuristic clustering methods to analyze the association of Bitcoin account addresses. 
    Reid\etc\quote{ReidH11}propose the first cluster method for re-identification, named multi-input heuristic, which assumes that the input addresses of a particular transaction are possessed by the same entity. 
    Then, Androulaki\etc\quote{AndroulakiKRSC13}propose the change address cluster method, which assumes that a new "change" address created by a transaction is likely controlled by the same entity that created the transaction.
	However, due to the anonymity of users in Web 3.0 and the need for real labeled data sets, the effectiveness of the clustering method cannot be verified. To solve these problems, Kappos\etc\quote{KapposYSRHM22}solve the common input heuristic's validation problem by combining the peeling chain's transaction principle. While Wu\etc\quote{WuLCHZZ22}extended the labeled dataset by the PU learning approach.
	
	In addition to traditional machine learning methods, some studies use graph neural networks to identify transaction entities in Web 3.0. 
    Shen\etc\quote{abs-2104-06559}construct the account interaction graphs using Ethereum and EOSIO data and propose an end-to-end graph convolution network model to identify different categories of accounts or bots. 
    Zhou\etc\quote{ZhouHCWSX22}use graph neural network technology to convert the transaction de-anonymization problem in the Ethereum platform into a subgraph classification problem, which improved the accuracy of identifying anonymous accounts.
	\begin{table*}[t]
		\caption{Research of Situation Awareness on AI}
		\label{EXISTING METHODS}
		\centering
		\small
		\renewcommand\arraystretch{1.2}
        \resizebox{\textwidth}{!}{
		\begin{tabular}{c|c|c|c|l}
			\toprule
			\textbf{Subjects} & \textbf{Refs.} & \textbf{AI Methods} & \textbf{Utilized Feature} & \multicolumn{1}{c}{\textbf{Web3.0 Tasks}} \\
			\midrule
			\multirow{4}[8]{*}{\makecell[c]{Transaction \\entity \\recognition }} & \cite{ReidH11}   & Clustering & Transaction inputs, amount and time & Transaction address association analysis \\
			\cmidrule{2-5}          & \cite{abs-2104-06559}   & GCN   & Relationship between account and contract & Transaction account type identification \\
			\cmidrule{2-5}          & \cite{ZhouHCWSX22}    & GNN   & Transaction amount and time & Transaction account identification \\
			\cmidrule{2-5}          & \cite{WuLCHZZ22}    & PUL   & Transactions order and amount & Transaction data labeling \\
			\midrule
			\multirow{3}[6]{*}{\makecell[c]{Malicious \\transaction \\identification }} & \cite{ChenPLLXZ21}  & GCN & Transactions between accounts & Phishing Scams Detection \\
			\cmidrule{2-5}          & \cite{WuYLYCCZ22}   & SVM   & Transaction amount and timestamp & Phishing Scams Detection \\
			\cmidrule{2-5}          & \cite{ChenZCNZZ18}  & XGBoost & Operation codes of the smart contracts & Ponzi contract identification \\
			\midrule
			\multirow{5}[10]{*}{\makecell[c]{Network \\behavior \\perception }} & \cite{jinpengTIFS}   & GCN   & Graph representation of traffic interraction features & Dapp Access Behavior Identification \\
			\cmidrule{2-5}          & \cite{yitingTIFS}  & RF & Cumulative downlink packet length & Web page access behavior recognition \\
			\cmidrule{2-5}          & \cite{mingweiTIFS}  & MC & State transition in SSL/TLS handshake & Web3.0 application classification \\
			\cmidrule{2-5}          & \cite{zhenboIWQOS}  & CNN   & Sequence of unidirectional burst lengths & Web3.0 website identification  \\
			\cmidrule{2-5}          & \cite{jinpengIWQOS}  & CNN   & Packet Round-Trip Time  & Video quality detection for network users \\
			\bottomrule
		\end{tabular}%
        }
		\label{tab:addlabel}%
	\end{table*}%
 
	\textbf{Malicious transaction identification}. Malicious asset trading behavior will have a bad impact on Web 3.0 ecosystem, and artificial intelligence technology can effectively improve the accuracy of identifying malicious trading behavior. 
	Common malicious transactions in Web 3.0 include Ponzi schemes, phishing websites, money laundering, etc. 
    In research on identifying Ponzi schemes, Chen\etc\quote{ChenZCNZZ18}extract features from user accounts and operation codes of the smart contracts and then built a classification model through the XGBoost algorithm to detect Ponzi schemes implemented as smart contracts. 
    For phishing websites detection, researchers extract features from the labeled transaction dataset that can mark phishing addresses and then use machine learning \cite{WuYLYCCZ22} or GCN \cite{ChenPLLXZ21} to transform the phishing site detection task into a classification problem based on the specific structure of the transaction features. 
    In terms of identifying money laundering transactions, GNNs have a significant advantage in analyzing graph structure-based transaction data \cite{abs-1908-02591}. And by enhancing the edge features of the trading graph \cite{abs-1906-05546}, GCN can also be used to identify money laundering accounts.
	
	\textbf{Network behavior perception.} Since many Web 3.0 applications use encrypted communication protocols such as SSL/TLS, most of the behavior traffic in the network appears in the form of ciphertext, which makes the key information contained in the plaintext invisible to regulators. AI has been widely used in network behavior perception of Web 3.0, including website fingerprinting \cite{zhenboIWQOS, yitingTIFS}, application traffic classification \cite{mingweiTIFS, jinpengTIFS} and video traffic classification \cite{jinpengIWQOS}.
 
	In constructing web page fingerprints, Shen\etc\quote{yitingTIFS}construct a web fingerprint classifier using random forest algorithm to identify web traffic by extracting packet length features. To fully extract the traffic features, \cite{zhenboIWQOS} trains a fine-grained website fingerprint classifier with CNN to achieve better results. In addition to being used to train website fingerprint classifiers, CNNs are also used to train video traffic classifiers to detect network video quality \cite{jinpengIWQOS}.
	In terms of application traffic classification, Shen\etc\quote{mingweiTIFS}use a second-order Markov model to construct a web application classifier to achieve the classification of encrypted web application traffic. Based on this, GNNs are used to train classifiers based on traffic interaction graphs \cite{jinpengTIFS}. 
	
	%\begin{figure}[htbp]
		%\small
		%\centering
		%\includegraphics[width=\linewidth]{images/WP.pdf}
		%\caption{Building Webpage Fingerprinting using RF \cite{yitingTIFS}}\label{WP}
	%\end{figure}

	\subsection{Summary and Lessons Learned}
	
	In this section, we introduce existing studies focusing on AI technology applied in Web 3.0 ecological management, including incentive mechanisms, content management, and situation awareness. The Web 3.0 ecosystem is subject to various anomalies in its operation. For example, users may create vulgar digital works to circulate, leading to ecosystem pollution. Due to the greater autonomy and anonymity of Web 3.0 users, it is costly and inefficient to detect these user anomalies manually. Artificial intelligence technology plays a huge role in assisting community administrators with anomaly detection and decision-making. However, because of the variety of unforeseen anomalies that can occur in the ecosystem, it is relatively rudimentary to train AI models to deal with these anomalies using labeled data.
	
	We summarize that data privacy breaches, poor model generalization, and lack of labeled data are three main dilemmas of Web 3.0 ecological management based on AI technology. Firstly, using privacy-preserving techniques such as federal learning, homomorphic encryption, and differential privacy can effectively avoid the privacy leakage problem in joint data training. Secondly, due to the dynamics and diversity of abnormal community behavior, the model should be lightly trained based on a small amount of anomalous behavior data, allowing the features and parameters of the model to be optimized and fine-tuned promptly. Lastly, combining unsupervised learning and reinforcement learning, which are less dependent on label data, can effectively improve the performance of the auxiliary management model.

 	\section{Application Layer} \label{sec:app}

	Web 3.0 is widely applied in finance, healthcare and game entertainment. In this section, we illustrate some applications and discuss the current situation of these applications and how to integrate with AI technology to provide more solid support for Web 3.0.

	\begin{table*}[t]
		\centering
		\caption{Research on AI-based Applications in Web 3.0}
        \renewcommand\arraystretch{1.2}
		\resizebox{\textwidth}{!}{
			\begin{tabular}{c|c|c|l|l}
				\toprule
				\textbf{Subject} & \textbf{Ref.}  & \textbf{AI Methods} & \multicolumn{1}{c|}{\textbf{Solutions}} & \multicolumn{1}{c}{\textbf{Web 3.0 tasks}} \\ 
				\midrule
				\multirow{3}[6]{*}{Finance} &  %\quote{LIU2021101755} & RNN, LSTM & Learning the characteristics of BTC network & Studying the trend of BTC \\
				%\cmidrule{2-5}          & \quote{CHEN2020112395} & SVM, LSTM & Using high-dimensional features & Predicting BTC price \\
				%\cmidrule{2-5}          &
                \quote{9534324} & DL    & AI-based systematic modular framework  & Detecting smart contract vulnerabilities \\
				\cmidrule{2-5}          & \quote{DBLP:journals/corr/abs-1811-06632} & LSTM  & Applying short and long term memory model & Learn vulnerabilities in sequence \\
				\cmidrule{2-5}          & \quote{10.5555/3491440.3491894} & GNNs  & Using Graph neural networks for detection & Smart contract vulnerability analysis \\
				\midrule
				\multirow{4}[8]{*}{Metaverse} & \quote{10.1145/3240508.3243653} & DRL   & Visual deep learning & Novel virtual environment establishment \\
				\cmidrule{2-5}          & \quote{9049708} & FL    & Federated learning based mobile edge computing & Proving computational effiency of AR applications \\
				%\cmidrule{2-5}          & \quote{9409764} & FL    & Federated learning-based DT framework & Multiple city DTs \\
				\cmidrule{2-5}          & \quote{10.1145/2134203.2134206} & RL    & Train virtual characters to move participants & Precomputing avatar behavior \\
				%\cmidrule{2-5}          & \quote{5345847} & RL    & A governance team based on RL & NPC team creation \\
				\cmidrule{2-5}          & \quote{DBLP:conf/chi/NakanoHIUK22} & CNNs  & Overlay food segmentation image inferred by CNNs & Improve the presence of users eating in metaverse \\
				%\cmidrule{2-5}          & \quote{7293159} & RL    & A metamorphic test mechanism & Mobile strategies in artificial games \\
				%\cmidrule{2-5}          & \quote{article} & DRL   & Combination of RL and deep network & AI agents in real-time combat games \\
				%\cmidrule{2-5}          & \quote{9314886} & CNNs, DQN & Combination of RL and supervised learning & AI agents in RTS games \\
				\midrule
				\multirow{2}[4]{*}{Healthcare} & \quote{9500397} & ANN   & AI-enabled and Blockchain-driven & Medical Healthcare System for COVID-19 \\
				%\cmidrule{2-5}          & \quote{9832978} & XAI   & Fusing Blockchain and explainable AI & Metaverse enabled telesurgery scheme \\
				\cmidrule{2-5}          & \quote{9153758} & DCNNs & An intermediate fusion framework & Physical activity recognition \\
				%\cmidrule{2-5}          & \quote{9072104} & CNNs  & A distributed hierarchical deep learning system & Wearable devices based fall detection \\
				\bottomrule
			\end{tabular}%
		}
		\label{tab:addlabel}%
	\end{table*}%
 \subsection{Finance}
	
    Web 3.0 has a wide range of applications in the financial field. The integration of AI is mainly used to predict the price of digital assets and improve the efficiency and security of Defi, or smart contracts. We have already discussed digital asset price forecasting in Section 5.2. In this section, we will focus on vulnerability detection and efficiency improvement of Defi, or smart contracts.

	\textbf{DeFi and smart contract.} Over the years, more and more projects have focused on this field and gradually evolved the concept of Defi. Defi, in short, is to make full use of blockchain technology (including smart contracts, decentralized asset custody, etc.) to replace all the intermediary roles in traditional financial services by code, to maximize the efficiency of financial services and minimize the cost. We will discuss the current situation of existing applications and focus on some smart contracts based on AI technology.
    The introductory text \cite{schueffel2021defi} discusses the origins of DeFi and delineates DeFi characteristics from those of traditional finance. Several examples of DeFi applications are given, the disadvantages resulting from this paradigm are discussed, and an outlook is provided. For example, Uniswap v3 \cite{adams2021uniswap} is a noncustodial automated market maker implemented for the Ethereum Virtual Machine. MakerDao \cite{MakerDaoDai} is a decentralized, unbiased, collateral-backed cryptocurrency soft-pegged to the US Dollar.%Compound \cite{Compound} is an EVM-compatible protocol that enables supplying of crypto assets as collateral to borrow the base asset. Accounts can also earn interest by supplying the base asset to the protocol.
    
	The rapid development of AI technology has helped improve the security and efficiency of DEFI. Some researchers have applied it to the detection of smart contract vulnerabilities. %Huang\etc\quote{DBLP:journals/corr/abs-1807-01868}convert the bytecode of the vulnerability contract into RGB (red, green, blue) images. And the convolution neural network algorithm is used to train these images to obtain the vulnerability detection model. 
    For instance, Tann\etc\quote{DBLP:journals/corr/abs-1811-06632}apply the LSTM model to learn vulnerabilities in sequence. However, these methods do not consider the impact of local code vulnerabilities on the overall code, which reduces the interpretability of these methods. To solve the problem of interpretation, Zhang\etc\quote{10.5555/3491440.3491894}explore the use of graph neural networks for smart contract vulnerability detection. They construct a contract graph to represent the syntax and semantic structure of smart contract functions and propose a Time Message Propagation (TMP) network to detect vulnerabilities. While Yu\etc\quote{9534324}propose the first systematic modular framework for detecting smart contract vulnerabilities based on deep learning, called DeeSCVHounter, which focuses on two types of smart contract vulnerabilities: reentry and time dependence. Their main innovation is to propose a novel Vulnerability Candidate Slice (VCS) concept to help the model capture the key points of vulnerabilities.

	\subsection{Metaverse}
	
	As one of the important applications of Web 3.0, Metaverse covers many fields, including finance, games, healthcare, and so on. This paper mainly researches two aspects of the Metaverse: environment establishment and user's behavior. Other application areas such as finance and healthcare will be described in their respective sections.
	
	\textbf{Environment establishment.} The users of the Metaverse, objects, or transactions in the physical world interact with the Metaverse, constantly developing and persistently representing the structure, behavior, and context of unique physical assets in the virtual world. With the breakthrough of digital transformation, the latest trend in each industry is to build digital twins, and the ultimate goal is to use them throughout the asset lifecycle through real-time data.
	
	The virtual world of the Metaverse has produced a large number of data, which makes the digital twin based on deep learning crucial. Aiming at the shortcomings of existing works such as small scenes or limited interaction with objects,  Lai\etc\quote{10.1145/3240508.3243653}propose a novel visual depth learning virtual environment to provide large-scale and diversified indoor and outdoor scenes. Augmented Reality (AR) devices can provide people with an immersive interactive experience, and their applications are sensitive to latency. Therefore, Chen\etc\quote{9049708}solve the computational efficiency of AR applications, low latency object recognition, and classification problems by combining the mobile edge computing paradigm with federated learning. In addition to the above, AI can help make virtual characters more intelligent. Kastanis\etc\quote{10.1145/2134203.2134206}propose a reinforcement learning method used to train virtual characters to move participants to the designated position.

	\textbf{User's behavior.} The user's behavior in the metaverse can be the behavior characteristics in the game or the simulation behavior in virtual reality (VR). %It is challenging to program Non-Player Characters (NPCs) to react towards social signals appropriately, which is important for immersive narrative games in VR. %Dobre\etc\quote{DBLP:journals/vr/DobreGP22}collaborate with two game studios to develop an immersive machine learning pipeline for detecting social engagement. They also collect data from participants-NPC interaction in VR, which is then annotated in the same immersive environment. 	
    In the early stage, Lugrin\etc\quote{lugrin2006ai}propose a method for the AI-based simulation of object behavior so that interactive narrative can feature the physical environment inhabited by the player character as an actor. The prototype based on the top of the Unreal Tournament game engine relies on a causal engine, which essentially bypasses the native physics engine to generate alternative consequences to player interventions. Then, to allow users to eat naturally in Virtual Environment (VE), Nakano\etc\quote{DBLP:conf/chi/NakanoHIUK22}propose Ukemochi to improve the presence of users eating in metaverse. Ukemochi seamlessly overlays a food segmentation image inferred by deep neural networks on a VE. Ukemochi can be used simultaneously as a VE created with the OpenVR API and can be easily deployed for the metaverse. Recently, some users protect their identities by arbitrarily changing their avatars. However, Meng\etc\quote{DeAnonymizationAttacksOnMetaverse}come up with a way to de-anonymize fake VR avatars called AvatarHunter. It achieves de-anonymization attacks by recording videos of multiple views in a VR scene, collecting the gait information of the victim’s avatar and preserving the avatar’s motion characteristics.

	\subsection{Healthcare}
	
	The applications of Web 3.0 in healthcare are mainly in the fields of Electronic Health Record (EHR) management. The EHR management in Web 3.0 integrates blockchain technology and AI, and better protects the privacy and security of patient data in medical services than before.

\begin{figure}[htbp]
		\centering
		\includegraphics[width=0.9\linewidth=0.8]{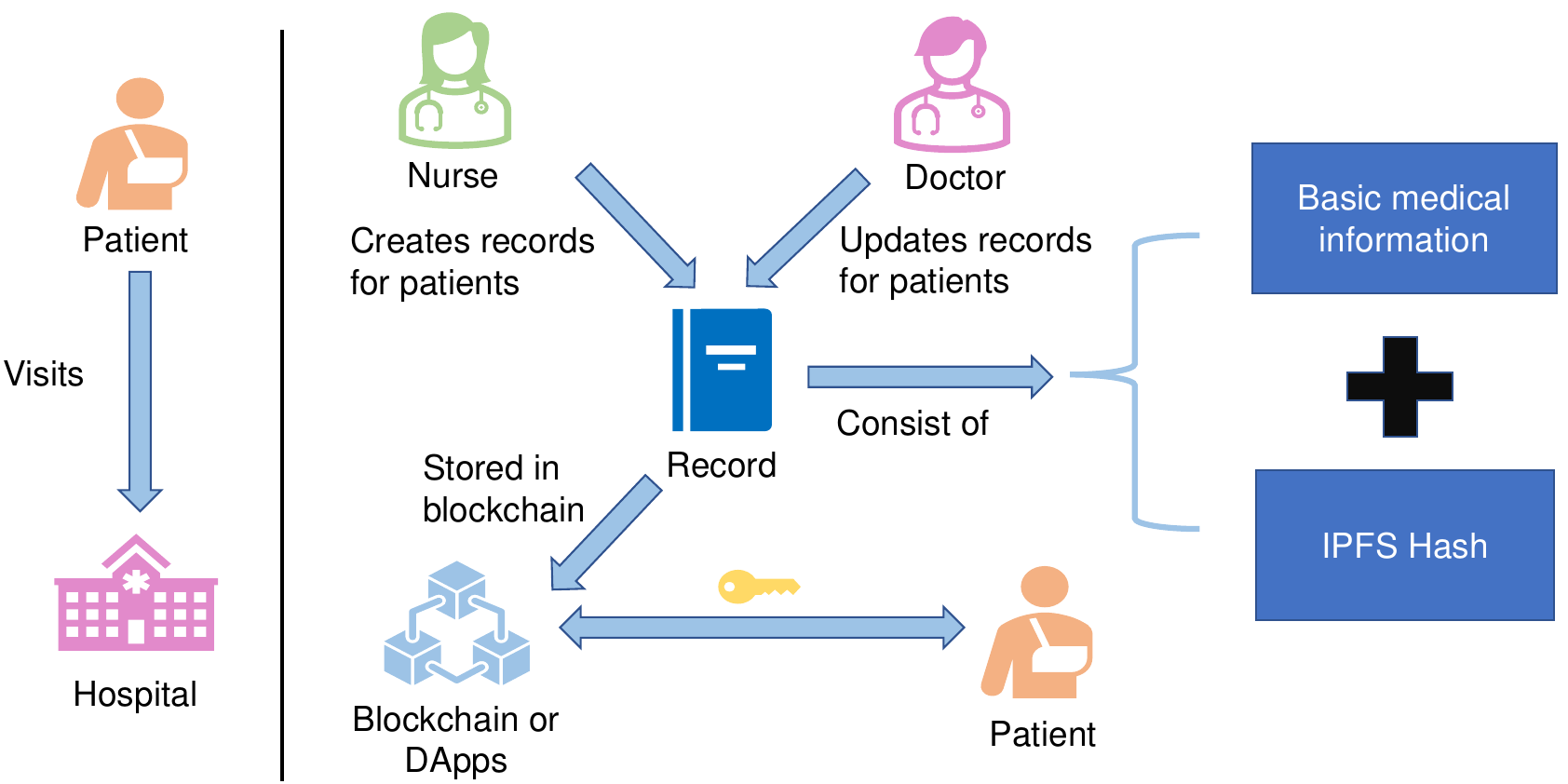}
		\caption{The framework of EHR management system based on Blockchain} \label{fig:EHRfigure}
	\end{figure}

	\textbf{EHR management.} The records in the traditional EHR management system are stored on a cloud server through the wireless communication channel, and there are risks of replay attacks, man-in-the-middle attacks, information leakage, and other security threats. Blockchain stores the data as a transaction with characteristics such as trust, and immutability, which also eliminates the intermediaries and a centralized dependence on transaction control. Fusing blockchain and EHR management is a better solution for the security threats and the framework of EHR management system based on blockchain as shown in Fig.~\ref{fig:EHRfigure}. One of the solutions given by Vora\etc\quote{8644088}is called BHEEM, which is a blockchain-based solution to store and efficiently transfer of EHRs. 
    %Also, Tanwar\etc\quote{TANWAR2020102407}propose a hyperledger fabric-based EHR system to maintain the security and privacy of patient’s data. 
    \\ \indent The above-described approaches in the literature failed to realize the significance of AI technology for EHR security, privacy, and accessibility.
    Mamoshina\etc\quote{mamoshina2018converging}propose a blockchain-based decentralized model that enables users to access their data in an AI-moderated healthcare data exchange. 
    Another study by Krittanawong\etc\quote{krittanawong2020integrating}on AI and blockchain integration can accelerate by greatly increasing the availability of data for AI training and development, able to share proprietary AI algorithms for generalization, decentralizing databases of different vendors or health systems, and incentivizing solutions that improve outcomes over those that do not, the integration of blockchain with AI could advance the goal of personalized cardiovascular medicine. 
    %Then, Witowski\etc\quote{witowski2021markit}propose MarkIt, a blockchain and AI based platform for collaborative annotation of medical imaging datasets. The platform enables radiologists to collaboratively annotate Digital Imaging and Communications in Medicine (DICOM) and non-DICOM images in order to create machine learning datasets and annotate them for classification and object detection tasks in an efficient manner. 
    For efficient contact tracing and monitoring of COVID-19 cases, Mistry\etc\quote{9500397}propose MedBlock, a novel AI-based and blockchain-driven EHR maintenance framework, which improves the efficiency of traditional EHR systems. %It seems promising to serve as a robust EHR maintenance system amidst the COVID-19 pandemic. 
    Some medical services in real life are also virtualized. By revealing the deep features of deep convolutional neural networks fused with traditional manual features, 
    Huynh-The\etc\quote{9153758}propose an intermediate fusion framework for human activity recognition (HAR) for supporting smart healthcare.

	\subsection{Summary and Lessons Learned}
	
	As mentioned earlier, we have discussed the details of AI and Web 3.0 applications in the fields of finance, metaverse, and healthcare, and the impact of AI on the performance of these applications. Although AI technology has not been fully integrated with these Web 3.0 applications, researchers have also tried to study AI blockchain technology and given some inspiring examples. %(such as the game project\quote{AIarena}introduced in the metaverse chapter).
	The integration of AI and blockchain, namely blockchain intelligence and intelligent blockchain, is worth exploring due to their close interaction. With the integration of AI and blockchain, distributed AI can process and execute the analysis or decision of trusted data without any support from a trusted third party. %Blockchain can provide various components for AI, including data sets, algorithms, and computing capabilities, through transactions in decentralized markets. 
    We believe that blockchain encourages AI to reach an unprecedented level in the context of various fields in Web 3.0.

    \section{Challenges and Future Research Directions} \label{sec:challenge}
	
	Although researchers have made substantial achievements in developing Web 3.0, there remain signiﬁcant challenges around the scalability of the infrastructure network, privacy protection, decentralized identity authentication, and abnormal transaction detection of digital assets, etc. At the same time, unprecedented study opportunities are also provided to develop innovative approaches to tackle these challenges.
	
	\subsection{Infrastructure layer}
	
	The challenges at the infrastructure layer can be summarized in two aspects. One is the performance challenge, namely the scalability problem, and the other is the challenge of security and privacy. The details are introduced as follows.
	
	\textbf{Scalability.} Scalability in Web 3.0 refers to the ability of a public blockchain system to process transactions, which is the key challenge faced by the Web 3.0 infrastructure layer and the primary obstacle that today's public blockchain systems are difficult to apply to practical scenarios. 
    %The reason is that as the number of nodes increases, the transaction will incur huge overhead to reach consensus at all nodes.
	The specifics of the scalability problem are as follows. (i) Transactions involve blockchain-related operations, which brings large processing costs; %At the same time, a large number of transactions significantly increase the resource overhead of data collection devices, which brings challenges to applications with high energy requirements (e.g. UAV); 
    (ii) On-chain data needs to be backed up at each node, which brings huge storage overhead, and a considerable number of nodes cannot afford this overhead; (iii) Due to the excessive amount of data in the network, Web 3.0 applications have certain data congestion and transmission delay compared to centralized applications, which is difficult to apply to real-time systems.
	
	As mentioned above, AI can make the system intelligent through learning algorithms, and minimize the system overhead by optimizing node behavior, system resource allocation, and other strategies. AI also plays a role in assisting optimization in scenarios such as blockchain sharding and cross-chain, effectively alleviating scalability problems. At the same time, it predicts the status of nodes to improve the security and availability of the system. 
	
	\textbf{Security.} In Web 3.0, the security problem is also worth attention. The security problems of the infrastructure layer mainly include (i) intrusion and attack by malicious entities; (ii) privacy of user data. (Other aspects like contract security, content security, etc., are outside the scope of this section.) This poses some challenges to the design of the infrastructure layer.	
	AI can solve the above problems to a large extent. The learning algorithm can be used to predict reliable nodes in the network, detect abnormal behaviors and identify intrusion behaviors of malicious entities. However, the model needs to use the user's data when training, and the current technology cannot provide strong privacy protection under the condition of guaranteeing performance. This requires the maturity of related privacy protection technologies, such as differential privacy, secure multi-party computation, etc.

	\subsection {Interface Layer}

    The challenges of the interface layer can be illustrated from two aspects, digital identity, and digital assets. The main problems in digital identity are the usability of the identity system. The main problems in digital assets are that the quality of AI-generated content still needs to be improved.

    \textbf {Digital ID.}
    At present, the high threshold of the identity system is one of the major problems. The public-private key system is currently the most widely used. Users need to remember the long and complex private key. Once forgotten, it cannot be retrieved. And users may also manage multiple addresses at the same time, which greatly increases the management cost. At present, the possible development direction is the private key recovery technology based on social relations and non-password technology. These methods will greatly reduce the cost of user identity management.

    Biometric authentication and behavioral authentication have shown their prospects as a means of identity verification. However, these methods are vulnerable to various types of attacks, including malware, imitation, simulation, deception, replay, statistics, algorithms, and robot attacks. The combination of multiple types of biometrics (i.e. multi-mode authentication) can improve security and provide more reliable authentication. Many studies have shown that multimodal biometric methods have advantages over single biometric methods.

    \textbf{Digital assets.}
    There are several challenges in the current artificial intelligence (AI) technology used to generate images. 
    It is difficult for AI to correctly depict spatial and physical relationships in generating images. For example, almost all AI can't draw a mirror well, because the image inside and outside the mirror needs optical knowledge, and the AI model is based on statistics and does not understand optics. Similarly, the geometry of AI painting is relatively bad. A typical example is that you will find that the wheels painted by AI are not too round. Another example is that the details of transparent glass glasses painted by AI are not correct.
    AI painting is independent of each other. It is difficult for AI to draw a complete set of works, such as a storytelling comic book with complex character relationships.
	
	\subsection{Management Layer}
	
	Although AI technologies can help Web 3.0 administrators better maintain order in their communities, there are still significant challenges as the Web 3.0 ecosystem matures and evolve. These challenges are specific to data set construction, personalized incentive mechanisms, and the performance of AI algorithms. Meanwhile, these challenges also point the way for the development of management technologies.

		\textbf{Incentive mechanism.}
	The incentive mechanism should unite benefit-sharing members in the blockchain-based trustless environment, and encourage them to effectively interact and collaborate around common goals according to their information, resources, goals, and risk appetite. 
    %The design and research of incentive mechanisms should consider the characteristics and needs of different types of members. 
    The incentives in Web 3.0 can be divided into transferable incentives and non-transferable incentives. Transferable incentives are mainly economic incentives in the form of fungible tokens or nonfungible tokens (NFTs). 
    Nontransferable incentives mainly refer to noneconomic incentives, including reputation, belief, and knowledge~\cite{weyl2022decentralized}. 
    How to combine various incentive measures and design appropriate mechanisms to meet the specific needs of each type of member and the common goal of the cooperation is a major challenge for incentive mechanisms in Web 3.0. 
    When designing an appropriate incentive mechanism for users, measurability, and personalization are particularly important, because the user's personal needs and optimization objectives largely affect the effectiveness of the incentive mechanism, and the needs and objectives vary from person to person. Therefore, the mixed incentive mechanism, which is mainly based on transferable incentives and supplemented by non-transferable incentives, can be used to quantify human contributions to the cooperation task, and then allocate reputation, certificates, tokens, governance rights, etc.

	\textbf{Content management.}
	Although content management has made significant progress in recent years, there are still many open issues worth exploring. The common seesaw nature between the generation method and the forensics method makes the Deepfake detection face many problems. %The challenges and possible future development directions in this field.	
	(i) %Generalization. The ability of a model trained by a specific forgery method to resist another unknown model. 
    The booming media manipulation techniques such as deepfake can generate more and more authentic and diversified forged data, and the potential forged types are usually unknown in real scenes. So media forensics algorithm should have good generalization ability.	
	(ii) %Robustness. The ability of the detector to cope with different scenarios. 
    The multimedia data such as images and videos spread on the network often undergo some post-processing, %such as compression, resizing, Gaussian blur, etc. This processing may destroy the details in the original data, 
    which makes the performance of the forensics model decline, which brings serious challenges to forgery detection. In addition, malicious users may deliberately impose invisible disturbance on forged data to deceive detection tools. Therefore, manipulation detectors should be robust to common distortion algorithms. 
    %\emph{3) Multi-modal information:} With the vigorous development of tampering technology, the performance of single-mode detection is limited. In contrast, different modes can jointly detect subtotal forgery artifacts, such as audio-visual joint detection.	
	%Model attribution. Most existing forensics methods usually focus on binary classification, namely true/false. However, attributing the manipulated images/videos to the model that generated them is also of great significance for legal accountability and intellectual property protection~\cite{girish2021towards}. The model attribution problem of Deepfake detection is far from being fully studied and solved. 
	(iii) %Interpretability. 
    Although DNN has strong discrimination ability, it lacks interpretability, so the judgment based on the neural network is difficult to accept in court and other serious occasions. However, the popular trend of deepfake detection methods is still based on deep learning, so it is important to improve the interpretability and credibility of these methods.

	\textbf{Situation awarness.} 
	Digital asset transaction dataset is a crucial component for situation awareness, as high-quality datasets can play an essential role in training. %e.g., helping to train accurate classifiers for supervised learning, and in validation, e.g., helping to evaluate the performance of abnormal transaction recognition methods, no matter whatever technique is used.
	%Digital Asset Transaction Dataset Construction.
    The construction of digital asset transaction datasets is a significant challenge in the current supervision of digital asset transactions.
	It is difficult to obtain digital asset transaction datasets with undisputed ground truth. Currently, two main methods are used in the literature to construct ground truth. We refer to a straightforward way to use an interface provided by a third-party service website to obtain the labeled data directly. This method is relatively simple, but it is often limited by the access restrictions of the third-party website, making it inefficient to obtain. At the same time, the labeled data provided by the third-party website has an intense lag time, and thus the researchers cannot access the latest labeled data promptly. Another method is for researchers to create digital asset transactions directly, allowing direct access to the data tags, but this method is extremely costly.
	
	To break through the dilemma between timeliness and low-cost, auto-labeling tools are highly desirable to automatically collect accurate and adequate data samples for a given analysis purpose. Since ensuring sufficient timeliness can be time-consuming and costly in terms of human efforts, crowdsourcing emerges as a promising approach to reduce the complexity, where multiple researchers or volunteers are involved in completing the same data collection and labeling tasks together. There are several issues to be further addressed, such as uniform standards for judging data quality, cost equalization for labeling data, and distribution of benefits from labeled data.

	\subsection{Application Layer}
	We have described some frontier applications of Web 3.0, covering the fields of finance, metaverse, and healthcare, and given some examples of the combination of Web 3.0 and AI. Then we will continue to describe the technical challenges and future research directions in these three fields.
	
	\textbf{Finance.}
	Although blockchain and smart contract research have made significant progress, blockchain research also faces a well-known trilemma, which makes it difficult to create a decentralized, scalable, and secure blockchain. At the same time, smart contracts at this stage are usually simple and automated contracts, which can only passively respond to predefined rules but cannot actively adapt to complex and dynamic environments. %Therefore, improving the intelligence level of smart contracts have attracted strong research interest, and this paper also gives some insights into this regard.	
	At present, artificial intelligence technology has not been widely used in the financial field of Web 3.0. %For the price prediction of cryptocurrency, we believe that more AI technology can be applied to improve the prediction success rate in the future. At present, blockchain and smart contracts are booming with diversity, and price prediction research for each cryptocurrency also needs to be carried out. 
    Because DeFi itself cannot be separated from smart contracts, if its application in the financial field wants to be further improved to gain more users' attention, AI technology is a good solution to the efficiency of smart contracts and privacy protection issues. AI smart contracts and blockchain are complementary. We believe AI smart contracts in the future can optimize energy consumption and improve mining efficiency, improve the scalability of blockchain, and be used to detect fraud.
	
	\textbf{Metaverse.}
	AI and blockchain are the basic technologies of the metaverse. The breakthrough of artificial intelligence technology, especially deep learning, has made great progress in the academic and industrial circles in the automatic operation and design of the metaverse.
	However, the existing deep learning model is usually very deep and has a large number of parameters, which brings a heavy burden to mobile devices with limited resources to deploy learning-based applications. Although the current AI technology is only at the stage when people tell machines to do specific tasks, rather than let machines learn automatic learning. Most learning tasks are only applicable to closed static environments with poor robustness and poor interpretability. %and cannot meet the requirements of usability, robustness, interpretability, and adaptability in an open dynamic environment. 
    Secondly, blockchain-related issues, such as whether the existing real-world NFT platform can adapt to the high transaction volume in the metaverse, and whether the metaverse needs a new blockchain platform and a new consensus mechanism, all of which deserve our deep consideration. In addition, %about the security and privacy of the metaverse, from the perspective of the metaverse company, developers, and users of the metaverse, 
    a natural problem is how to ensure the security and privacy in the metaverse, which may mean the violation of their privacy, potential identity theft and other types of fraud. In the future, solving these problems of the metaverse is crucial for its development.
	
	\textbf{Healthcare.}
	At present, some Web 3.0 applications in the medical field combine AI technology and blockchain. The current challenges faced by Web 3.0 applications in the medical field are mainly two aspects. On the one hand, the privacy of user data in the EHR management system cannot be adequately guaranteed, especially in the case of the COVID-19 pandemic. On the other hand, medical services in the metaverse are still in the initial stage. AI technology has provided a great promotion for virtual reality, But at present, there are many medical services in real life that cannot be realized in the metaverse. We believe that in future work, researchers can pay attention to the flexible application of machine learning in the EHR management system (especially FL) to better protect the privacy of users' medical records. In addition, we also expect more medical services in the metaverse (such as online consultation, online surgery, etc.) to emerge in the future.
	
	\section{Conclusion} \label{sec:conclu}
	
	This paper surveys the use of AI in Web 3.0. We begin by reviewing the development history of the web and the different perspectives on Web 3.0 from academic and business communities, then give our understanding of Web 3.0 and construct an architecture of the Web 3.0 ecosystem. 
 
	We divide the ecosystem of Web 3.0 into four layers: data management, value cycle, ecological governance, and application scenarios. The key focus of data management research is the use of intelligent strategies based on learning algorithms to enhance data management and resource allocation efficiency. The value cycle research focuses on identifying and authenticating digital identities, as well as pricing and transaction circulation of digital assets. Ecological governance has two main aspects: managing user behavior, such as abnormal transaction behavior and malicious traffic attacks, and managing user-generated content, including detecting bad or false information. The last section of the article mainly introduces the current AI applications in Web 3.0 from the field of finance, healthcare, and the metaverse. Finally, we also discuss the challenges and future directions for the application of AI in Web 3.0 and provide an outlook for its future development.
	
	All in all, with the rapid development of Web 3.0, the integration of AI technology will play an increasingly important role. We hope that this survey can provide a comprehensive reference for scholars who are interested in the application of AI technology in Web 3.0 and can also provide useful guidance for industry readers who are engaged in innovation in related fields. Further development on Web 3.0 needs collaboration from both academia and industries to strive for an open, fair and intelligent future Internet.

\nocite{}
\bibliographystyle{IEEEtran}
\bibliography{software}

\end{document}